%% file: main.tex
\documentclass{article} 
\usepackage[final]{colm2025_conference}

\usepackage{graphicx,subcaption}
\usepackage[utf8]{inputenc} 
\usepackage[T1]{fontenc}    
\usepackage{hyperref}       
\usepackage{url}            
\usepackage{booktabs}       
\usepackage{amsfonts}       
\usepackage{nicefrac}       
\usepackage{microtype}      

\hypersetup{
    colorlinks,
    linkcolor={red!50!black},
    citecolor={blue!50!black},
    urlcolor={blue!80!black}
}

\input{preamble}

\usepackage{lineno}

\definecolor{darkblue}{rgb}{0, 0, 0.5}
\hypersetup{colorlinks=true, citecolor=darkblue, linkcolor=darkblue, urlcolor=darkblue}

\title{To Backtrack or Not to Backtrack: \\
When Sequential Search Limits Model Reasoning}

\author{Tian Qin\thanks{Correspondence to \texttt{tqin@g.harvard.edu} \textsuperscript{\textdaggerdbl} Equal senior contributions.}\\
Harvard University 
\And
David Alvarez-Melis\textsuperscript{\textdaggerdbl} \\
Harvard University, Kempner Institute, MSR
\And
Samy Jelassi\textsuperscript{\textdaggerdbl} \\
Harvard University, Kempner Institute 
\And
Eran Malach\textsuperscript{\textdaggerdbl}\thanks{Currently at Apple.}  \\
Harvard University, Kempner Institute 
}

\begin{document}

\ifcolmsubmission
\linenumbers
\fi

\maketitle

\input{sections/abstract}
\input{sections/intro}

\input{sections/related}
\input{sections/method}
\input{sections/trace_study}

\input{sections/two_factor}
\input{sections/rl}
\input{sections/conclusion}
\clearpage

\section*{Acknowledgments}
We thank Core Francisco Park and Bingbin Liu for helpful discussions and feedback throughout the development of this work. TQ and DAM acknowledge support from the Kempner Institute, the Aramont Fellowship Fund, and the FAS Dean’s Competitive Fund for Promising Scholarship.

\bibliography{paperpile.bib}
\bibliographystyle{colm2025_conference}
\clearpage
\appendix
\input{sections/appendix/real_math}

\input{sections/appendix/related}
\input{sections/appendix/hyperparameter}

\input{sections/appendix/compute_flops}
\input{sections/appendix/majority_vote}
\input{sections/appendix/game_type}
\input{sections/appendix/cd_search_strat}
\input{sections/appendix/confusion_table}
\input{sections/appendix/training_curve}
\input{sections/appendix/grpo}
\input{sections/appendix/verbatim}

\end{document}

%% file: preamble.tex
\usepackage{microtype}
\usepackage{graphicx}
\usepackage{subcaption}
\usepackage{booktabs}
\usepackage[dvipsnames]{xcolor}
\usepackage{caption}
\usepackage{wrapfig}
\usepackage{graphics}
\usepackage{enumitem}
\usepackage{amsmath}
\usepackage{amssymb}
\usepackage{mathtools}
\usepackage{amsthm}
\usepackage[skip=10pt plus1pt, indent=0pt]{parskip}
\let\svthefootnote\thefootnote
\newcommand\freefootnote[1]{%
  \let\thefootnote\relax%
  \footnotetext{#1}%
  \let\thefootnote\svthefootnote%
}
\usepackage{multirow}
\usepackage{algorithm}
\usepackage{algpseudocode}
\usepackage{duckuments}
\usepackage{pifont}
\usepackage{tcolorbox}
\usepackage{fancyvrb}

\makeatletter
\let\blx@rerun@biber\relax
\makeatother

\makeatletter
\let\blx@rerun@biber\relax
\makeatother

%% file: sections/abstract.tex
\begin{abstract}
Recent advancements in large language models (LLMs) have significantly improved their reasoning abilities, particularly through techniques involving search and backtracking. Backtracking naturally scales test-time compute by enabling sequential, linearized exploration via long chain-of-thought (CoT) generation. 
However, this is not the only strategy for scaling test time-compute: parallel sampling with best-of-$n$ selection provides an alternative that generates diverse solutions simultaneously. Despite the growing adoption of sequential search, its advantages over parallel sampling---especially under a fixed compute budget---remain poorly understood.
In this paper, we systematically compare these two approaches on two challenging reasoning tasks: CountDown and Sudoku. 
Surprisingly, we find that sequential search underperforms parallel sampling on CountDown but outperforms it on Sudoku, suggesting that backtracking is not \textit{universally} beneficial.
We identify two factors that can cause backtracking to degrade performance: (1) training on fixed search traces can lock models intro suboptimal strategies, and (2) explicit CoT supervision can discourage `implicit` (non verbalized) reasoning. Extending our analysis to reinforcement learning (RL), we show that models with backtracking capabilities benefit significantly from RL fine-tuning, while models without backtracking see limited, mixed gains. 
Together, these findings challenge the assumption that backtracking universally enhances LLM reasoning, instead revealing a  complex interaction between task structure, training data, model scale, and learning paradigm.
\end{abstract}

%% file: sections/intro.tex
\section{Introduction}
\begin{figure}[t]
    \centering
    \includegraphics[width=1.0\linewidth]{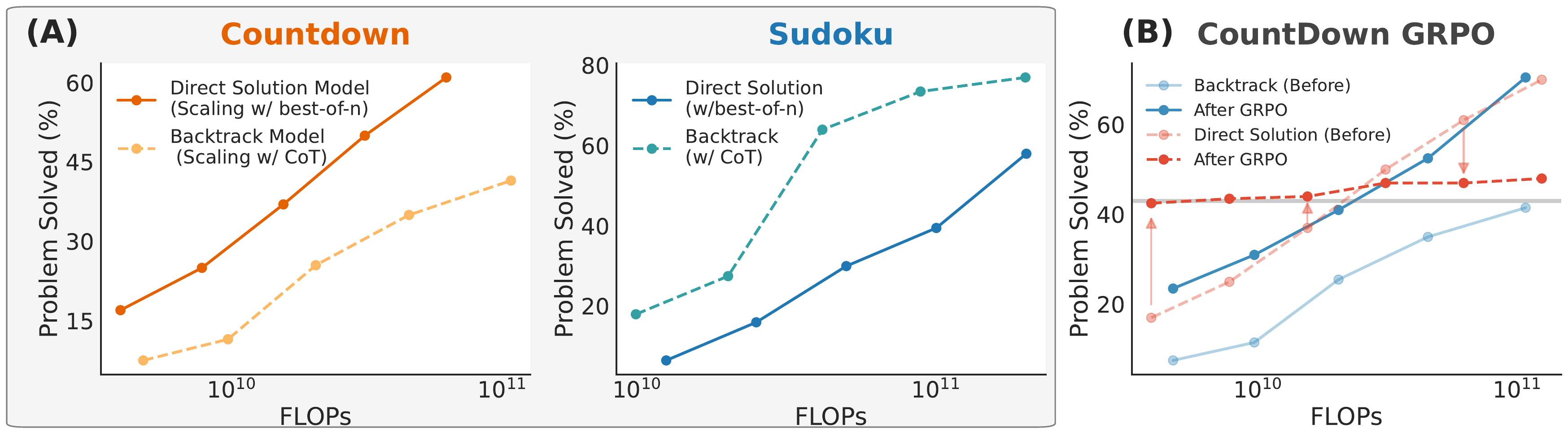}
    \caption{
    \textbf{
    Backtracking performance varies significantly with task type and the application of post-training reinforcement learning.
    }
    \textit{(A)} Training backtracking and direct solution models on CountDown and Sudoku reveals task-dependent performance: under equal test-time compute, backtracking (sequential search) underperforms direct solution with best-of-$n$ generation (parallel search) on CountDown, but outperforms it on Sudoku.
    \textit{(B)} Fine-tuning with GRPO consistently improves backtracking model performance across compute budgets, but has mixed effects on the direct solution model.
    }
    \label{fig:main_figure}
\end{figure}

Recent studies \citep{Kumar2024-se, Havrilla2024-bl} propose teaching LLMs to correct mistakes through \textit{backtracking}, enabling exploration of alternative solutions. Despite growing popularity \citep{DeepSeek-AI2025-vs, Muennighoff2025-ms}, it remains unclear whether correcting errors post-hoc via backtracking is ultimately more compute-efficient at test time than directly learning the correct solution.
Solving strategic games such as CountDown and Sudoku requires extensive exploration of different solution paths, making them ideal for analyzing the computational trade-offs of sequential versus parallel search.
In this work, we use these two games to conduct a controlled investigation to determine whether backtracking is an effective way to scale test-time compute.

There are two primary strategies to scale LLMs’ test-time compute: sequential autoregressive search (explicit backtracking within a chain-of-thought) and parallel sampling (generating multiple independent solutions and selecting the best with best-of-$n$). While sequential search allows the model to refine reasoning by learning from past mistakes, it comes at a cost: due to the attention mechanism, the FLOPs required to generate CoT grow quadratically with sequence length. Even when generating the same number of tokens, sequential search incurs more FLOPs than parallel sampling. To compare these two strategies, we train (i) \textbf{backtracking models} that learn from explicit search traces and use sequential search to solve hard problems, and (ii) \textbf{direct solution (i.e., no backtracking) models} that learn solely from correct solutions, using parallel search at test time. Equating test-time compute, we observe contrasting results (Fig.~\ref{fig:main_figure} \textit{A}): in CountDown, the backtracking model consistently \textbf{underperforms}, whereas in Sudoku, it consistently \textbf{outperforms} the direct solution model.

Through controlled experiments, we identify two reasons teaching backtracking can inadvertently degrade performance. First, explicit backtracking reasoning traces bias models toward prescribed search strategies, \textbf{limiting exploration} of potentially superior alternatives. In CountDown, the backtracking model closely mimics training search paths, while the direct solution model independently discovers more efficient strategies (Section \ref{sec:good_bad}). Second, detailed backtracking traces \textbf{encourage verbosity} (producing lengthy yet ineffective reasoning chains), while discouraging internal "thinking" (implicit reasoning without outputting CoT, Section \ref{sec:two_variations}). Beyond these factors, we demonstrate that model size and task-specific characteristics also impact the effectiveness of backtracking (Section \ref{sec:model_size}). Crucially, we show that our contrastive observation between Sudoku and Countdown \textbf{generalizes} to real-world tasks: such as math and science problem solving. We show that backtracking is \textit{not} always the most effective way to scale test-time compute (Appendix~\ref{appdx:real_math}) for general reasoning models.

Extending beyond supervised learning, we evaluate reinforcement learning (RL) with Group Relative Policy Optimization (GRPO) \citep{Shao2024-me}, uncovering novel interactions between backtracking capabilities and RL. We show that the backtracking model discovers new, effective search strategies through RL, achieving substantial performance improvements. Conversely, the direct solution model improves one-shot accuracy but loses effectiveness in parallel search, revealing a clear trade-off (Fig.~\ref{fig:main_figure} \textit{B}). This finding shifts our understanding of how backtracking influences a model’s potential to improve under RL, highlighting the unique advantage of teaching backtracking for long-term reasoning capabilities.

Our controlled study on two strategic games provides a nuanced understanding of when backtracking effectively scales test-time compute. Our main contributions are:

\begin{itemize}[itemsep=2pt,labelindent=2pt,topsep=0pt,parsep=0pt,partopsep=1pt, align=left, leftmargin=*]
    \item We use CountDown and Sudoku as controlled testbeds to examine whether backtracking enables efficient test-time scaling. Under a fixed compute budget, backtracking outperforms parallel search in Sudoku but underperforms in CountDown (Fig.~\ref{fig:main_figure} \textit{A}).
    
    \item We identify two key factors affecting backtracking efficacy: (1) \textbf{Prescribed search bias}: Training on detailed backtracking traces can unintentionally constrain models to suboptimal search strategies. (2) \textbf{Excessive verbosity:} Explicit backtracking traces encourage models to produce lengthy reasoning chains without improving reasoning ability.

    \item We demonstrate that reinforcement learning (GRPO) consistently enhances backtracking models by enabling discovery of novel solutions, whereas direct solution models experience mixed outcomes (Fig.~\ref{fig:main_figure} \textit{B}).
\end{itemize}

%% file: sections/related.tex
\section{Related Work}

\begin{figure}[t]
    \centering
    \includegraphics[width=0.9\linewidth]{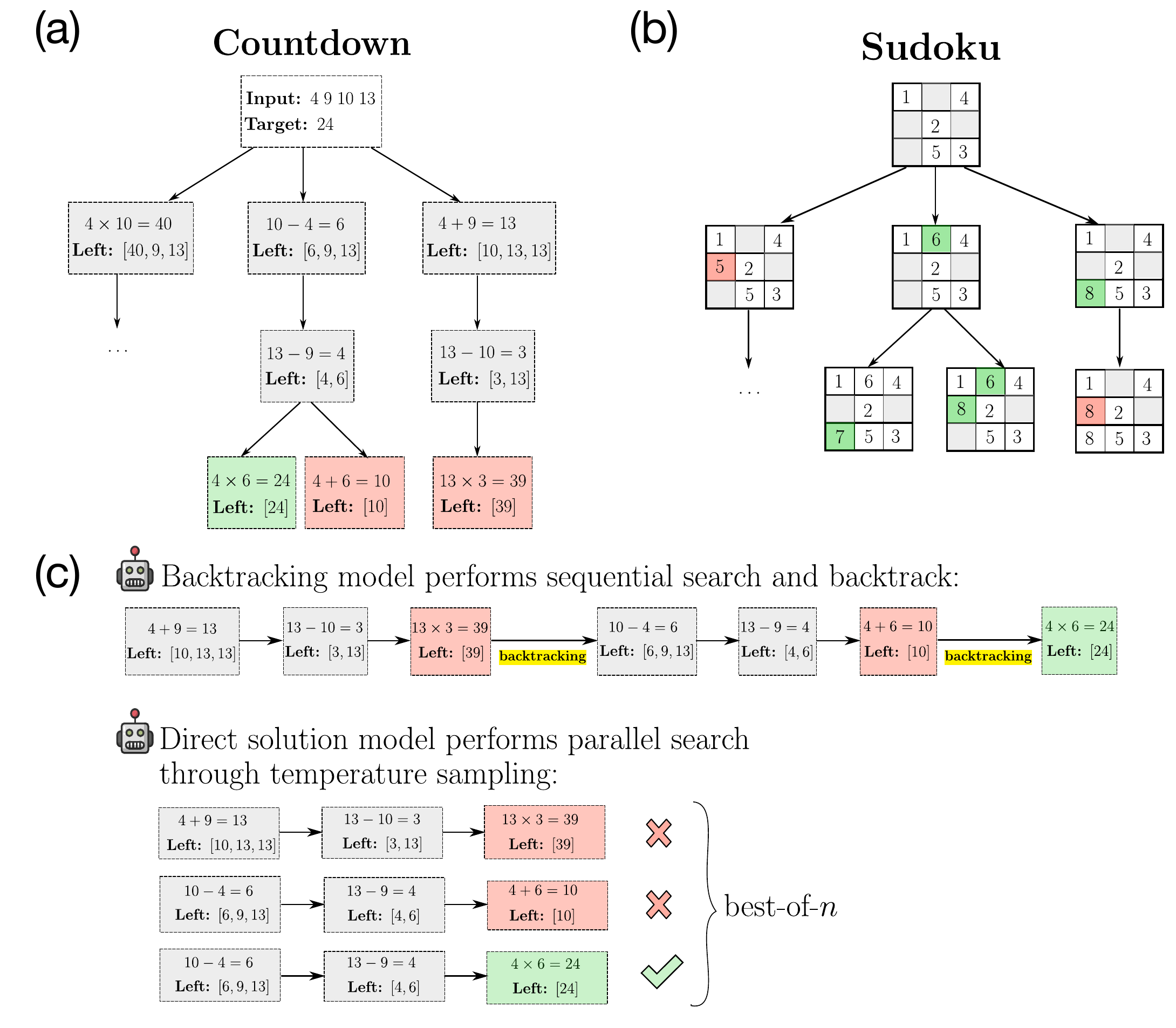}
    \caption{
    \textbf{Backtracking and direct solution for two different strategic games.}
    \textit{Panel (a, b):} Example the search tree for CountDown and Sudoku. Solving both games require extensive search in the solution space. 
    \textit{Panel (c):} The backtracking model is trained on the search traces generated by a Depth-First-Search (DFS) algorithm. At test time, the model performs sequential search. The direct solution model is trained on the correct solution only. At test time, the model performs parallel search through temperature sampling and takes best-of-$n$. 
    }
    \label{fig:setup_demo}
\end{figure}

See Appendix \ref{appdx:related} for an extensive review on related work. 
\paragraph{Scaling test-time compute .}
Prior work has explored scaling language model performance at test time through parallel or sequential search strategies. Parallel methods rely on independent sampling and selection via heuristics or reward models \citep{Brown2024-zj, Irvine2023-jv, Levi2024-wd, Xin2024-rj}, while sequential methods refine reasoning step by step using earlier outputs \citep{Hou2025-qz, Lee2025-tw}. Tree-based methods such as MCTS bridge the two and often incorporate process-level reward models to guide reasoning \citep{Wu2024-bj, Lightman2023-cg}. Our work contributes to this area by comparing sequential (backtracking) and parallel search under fixed compute budgets.

\paragraph{Self-correction and backtracking.}
Language models can be trained to self-correct through fine-tuning on revision data, synthetic augmentations, or reward-based learning \citep{Saunders2022-cr, Qu2024-rl, Welleck2022-jb}. Some approaches also introduce explicit search or separate correction modules to guide revision \citep{Yao2023-qu, Havrilla2024-bl}. We build on this line of work by studying backtracking as an implicit form of self-correction, analyzing when learning to backtrack helps or hinders reasoning.

\paragraph{Reinforcement learning for LLM reasoning.}Reinforcement learning has shown promise in enabling language models to autonomously discover reasoning strategies, including through simplified algorithms like GRPO \citep{Shao2024-me, DeepSeek-AI2025-vs}. While prior work has demonstrated strong results, it remains unclear which model properties enable successful RL-based reasoning \citep{Zelikman2022-yo, Kazemnejad2024-gw}. Our study addresses this gap by comparing how backtracking and no backtracking models respond to RL fine-tuning, revealing asymmetric benefits.

%% file: sections/method.tex
\section{Two strategic games: CountDown and Sudoku}
\subsection{CountDown}
\subsubsection{Game setup}
The Game of CountDown has been frequently used as a testbed to study and evaluate LLM reasoning \citep{Gandhi2024-tu, Gandhi2025-ac, Yao2023-fg}. In a CountDown game, the player is given a set of candidate numbers and a target number (restricted to integers). The goal is to reach the target by applying a sequence of arithmetic operations—addition, subtraction, multiplication, or division—using the candidate numbers. Each number must be used exactly once, and intermediate results can be reused in subsequent operations. 

To algorithmically solve CountDown, we can represent the problem as a search tree (Fig.~\ref{fig:setup_demo}\textit{a}). Each node in the search tree corresponds to a state defined by the current set of available numbers. At each step, the algorithm selects a pair of numbers from the set and applies one of the four operations, replacing the pair with the resulting value to create a new state. This process continues recursively until the target number is reached (correct leaf node) or all combinations are exhausted (wrong leaf node). In this work, we play the CountDown with four candidate numbers, and for each game, there are 1,152 possible search paths. 

\subsubsection{Data generation}
\label{sec:cd_data}
We generate \textbf{backtracking traces} with Depth First Search (DFS) with a sum-heuristic (\citet{Gandhi2024-tu}, further details in Appendix \ref{appdx:data_detail}). 
We generate a dataset of 500,000 CountDown questions, and the DFS search correctly solves 57\% of the questions. The backtracking trace is a serialized version of DFS, listing all the tree nodes visited in the order of DFS traversal. To construct the \textbf{direct solution} training data, we prune the backtracking traces to keep only the correct solution path. With the pruning approach, we remove the exploratory parts of the trace while preserving the answer format and scaffolding used in the backtracking model, to ensure a fair comparison. We also ensure that the direct solution model does not see more solved CountDown games, we include only the 285,000 questions (i.e., $500{,}000 \times 0.57$) that DFS successfully solves. We provide examples of both training data in Appendix \ref{appx:verbatim}.

\subsection{Sudoku}
\subsubsection{Game setup}
Sudoku is another prototypical strategic game used to study reasoning and search in LLMs \citep{Yao2023-fg, Long2023-cb}. In this work, we focus on hard $9 \times 9$ Sudoku boards, where only about 20 of the 81 cells are pre-filled, making the search space substantially larger (see Appendix~\ref{appdx:data_detail} for a description of Sudoku rules). To algorithmically solve Sudoku, we represent the problem as a search tree (Fig.~\ref{fig:setup_demo}\textit{b}). Each node corresponds to a partial board state, where some cells have been filled. At each step, the algorithm selects an unfilled cell and fills it with a candidate digit that satisfies Sudoku constraints in the current state. Each valid assignment creates a new child node representing the updated board. The process continues recursively until a complete, valid solution is reached (correct leaf node) or no valid moves remain (wrong leaf node). The depth of the tree corresponds to the number of empty cells, and the branching factor at each node depends on the number of unfilled cells as well as how many digits are valid for each unfilled cell.

\subsubsection{Data generation}
\label{sec:sudoku_data}
We follow the same procedure as CountDown to generate training data for both the backtracking and direct solution models. We use a DFS-based search algorithm, in combination with a Sudoku solver that applies seven common human strategies (e.g., naked singles, hidden pairs and etc, \citet{Papadimas2023-xd} ) to eliminate candidates for unfilled cells. At each node, we use the 7 strategies to eliminate candidates for unfilled cells, and then DFS chooses an unfilled cell, makes a guess and continues solving recursively. This process continues until the board is either solved or reaches a dead-end (i.e., an invalid state with no legal moves). We use a dataset of 3M Sudoku puzzles from \citep{Radcliffe2020-as}, and the combined DFS-solver approach successfully solves 98\% of them. Since DFS successfully solves nearly all puzzles, we train both models on 2.8M examples and reserve the last 200K for validation and testing. We provide further details on Sudoku training data generation in Appendix \ref{appdx:data_detail} and data examples in Appendix \ref{appx:verbatim}.

\subsection{Model and training}
We use Qwen2.5-style model architectures \citep{Yang2024-so} with RoPE positional encoding \citep{Su2021-qu} and Group Query Attention (GQA) \citep{Ainslie2023-wh}. To maximize parameter efficiency, we design custom tokenizers for both games, significantly reducing the size of the language modeling head. This allows us to train smaller models than prior work \citep{Gandhi2024-tu, Shah2024-ri} while maintaining comparable performance on both tasks. For CountDown, we use a 17M parameter model with a context length of 4096 tokens; for Sudoku, we use a 38M model with the same context length. See Appendix \ref{appdx:hparam_train} for model architecture and an exhaustive list of training hyperparameters. We train all models until validation loss converges (see Appendix \ref{appdx:training_curve}).

%% file: sections/trace_study.tex
\section{Empirical trade-offs of backtracking}
\begin{figure}[t]
    \centering
    \includegraphics[width=1.0\linewidth]{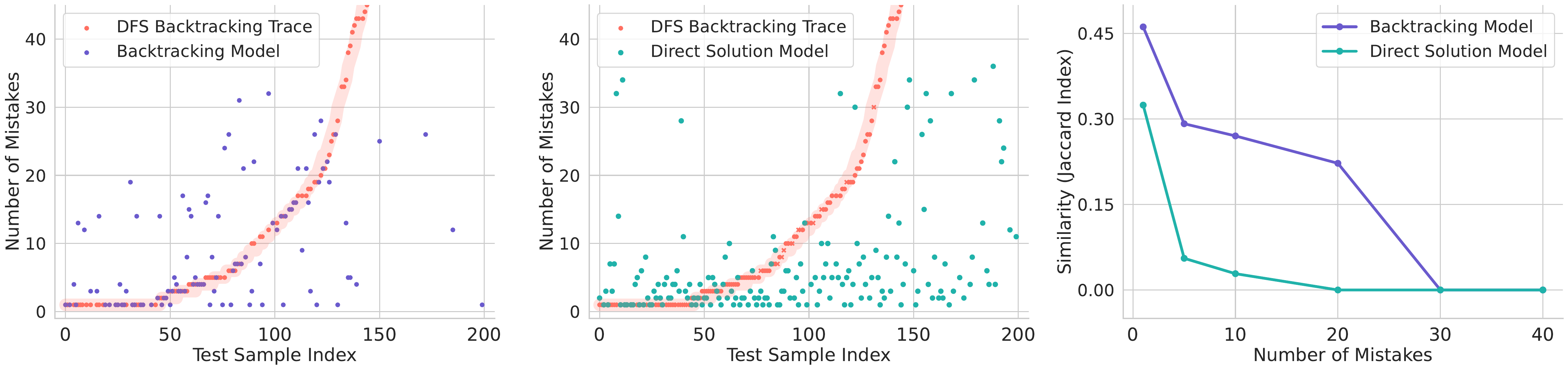}    
    \caption{
    \textbf{Backtracking and direct solution models implement different search strategies for CountDown.} For test questions that model solves correctly, we measure the number of mistakes made (i.e., wrong terminal nodes visited) before finding the correct solution. We sort the test questions by number of mistakes made by DFS. 
    \textit{Left:} Trained on DFS traces, the number of mistakes made by the backtracking model correlates with the DFS.
    \textit{Middle:} In contrast, the direct solution model solves a lot more problems with significantly fewer mistakes compared to DFS.  
    \textit{Right:} For a given number of mistakes made, we examine whether two models solve the same set of question as DFS. Direct solution model implements a search strategy significantly different from DFS. 
    }
    \label{fig:countdown_performance_correlation}
\end{figure}
We first demonstrate that backtracking models do not universally outperform the direct solution models (Section \ref{sec:main_result}) because backtracking models are restricted to learn a prescribed way of search (Section \ref{sec:good_bad}). We then identify two factors (Sections \ref{sec:two_variations}) showing how we might improve test-time scaling for backtracking models. 

\subsection{Backtracking is not always beneficial}
\label{sec:main_result}
\paragraph{Evaluation metrics.}
We evaluate model performances using solving accuracy on 200 unseen problems with binary scores (either correct or incorrect, no partial credits, see appendix \ref{appdx:data_detail}). We use FLOPs to compare inference costs (see Appendix \ref{appdx:compute_flops} for FLOPs computation). For the backtracking model, we allow models to autoregressively generate and measure how many problems the model finds the correct solution at various CoT lengths (ranging from 1024 to 4096 tokens). For the direct solution model, we generate $n$ solutions in parallel through temperature sampling at $T=0.7$, and examine whether the model has found the correct solution within $n$ attempts (i.e., best-of-$n$). Best-of-$n$ is a suitable choice in those two games, a case where solving the task is hard but verification is trivial. In general, our analysis applies to tasks where verification can be easily done with an external verifier at test-time. This is definitely not always the case, and we leave the study of problems where test-time verification is not as easy to future work. In those tasks, one might need to consider majority voting or other strategies. See Appendix \ref{appdx:majority_vote} for further discussions.

\paragraph{Results.}
In Fig.~\ref{fig:main_figure} \textit{A}, we observe distinct scaling behaviors for the two models. For both games, the direct solution model’s test accuracy scales \textit{linearly} with increased test-time compute (measured on a logarithmic scale). This scaling behavior indicates that through parallel sampling, the backtracking model generates diverse solutions that search through different solution paths. Conversely, the backtracking model exhibits \textit{sub-linear} scaling: Longer solution traces disproportionately yield smaller accuracy improvements. We attribute the sub-linear scaling to two causes. First, as reasoning chains become longer, the backtracking model might struggle to effectively track visited states and efficiently search through the solution space. Second, when models perform sequential search, the computation cost grows quadratically with CoT length (due to the attention mechanism, see Appendix \ref{appdx:compute_flops}), and this further makes backtracking model less effective for scaling up test time compute. Overall, for CountDown, the direct solution model consistently outperforms its backtracking counterpart. However, this trend is reversed in Sudoku, where the backtracking model consistently achieves higher accuracy.

\subsection{Backtracking model learns both the good and the bad}
\label{sec:good_bad}

When teaching a child to correct math mistakes, the child understands that the goal is the correct answer---not making and then fixing errors. Humans have meta-cognitive awareness that models lack. Models trained via next-token prediction simply imitate the traces they see, including making the mistake before fixing it. In CountDown, this poses a key limitation: the backtracking model learns to follow the specific search paths seen in training. While some tasks---like shortest path finding---have optimal strategies we can supervise directly (e.g., Dijkstra’s algorithm), most reasoning tasks, including CountDown, lack such guarantees. As a result, the model may be constrained by the inefficiencies in the backtracking data. In contrast, the direct solution model, trained only on correct answers, is free to discover more efficient strategies. In our subsequent analysis, we concretely show how the direct solution model successfully bypasses many inefficient search and backtracking steps learned by the backtracking model.

\subsubsection{Backtracking model finds the solution with fewer mistakes}
\paragraph{Measuring number of mistakes.}
We compare the number of mistakes made by: (1) DFS (used to generate backtracking data), (2) the backtracking model, and (3) the direct solution model. For DFS and the backtracking model, mistakes are counted as the number of incorrect terminal nodes explored before finding the correct solution. For the direct solution model, mistakes correspond to how many parallel samples ($n$ in best-of-$n$) are needed. \footnote{Mistakes are counted only for problems solved correctly by the model.}

\paragraph{Comparing search strategies.}
\label{sec:search_strat}
We sort the 200 test problems based on mistakes made by DFS and plot mistakes for both models. Fig.~\ref{fig:countdown_performance_correlation} \textit{left} compares DFS search and backtracking model. The number mistakes made by the backtracking model is correlated with the DFS backtracking trace. This observation is not surprising given that the backtracking model is trained on these traces. However, this result is interesting when we compare it against the direct solution model (Fig.~\ref{fig:countdown_performance_correlation} \textit{middle}). The direct solution model solves most problems within fewer than 10 attempts---far fewer compared to DFS or the backtracking model.
Fig.~\ref{fig:countdown_performance_correlation} \textit{right} quantifies these observations. Specifically, for a fixed mistake budget, we use Jaccard Index to measure whether the model solves a similar set of problems as DFS solves. The backtracking model closely mirrors DFS search (high set similarity), whereas the direct solution model diverges significantly (low set similarity). Together with superior performance of the direct solution model, we conclude that the direct solution model learns more efficient search strategies, avoiding unnecessary explorations of wrong paths.

\begin{figure}[t]
    \centering
        \includegraphics[width=1.0\linewidth]{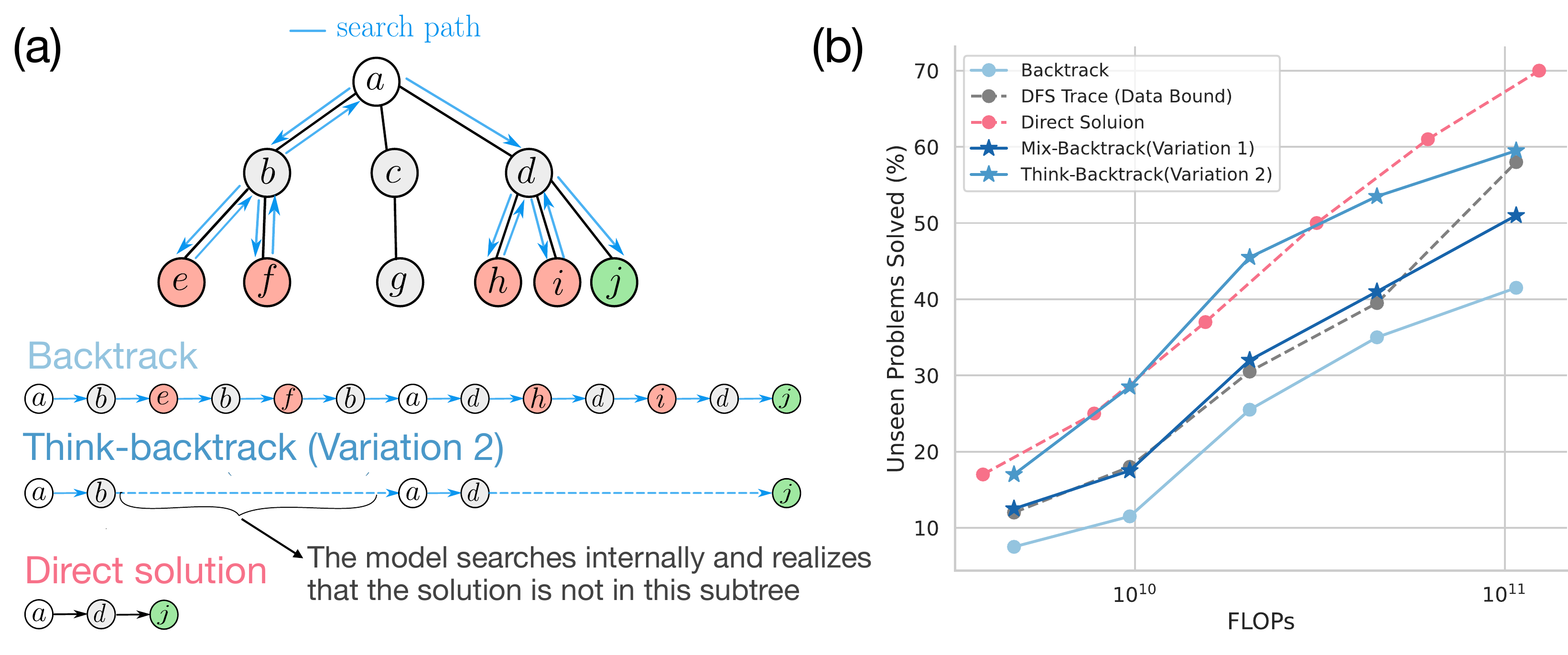}
        \caption{
        \textbf{Two different variations to improve backtracking model.}
        \textit{(a).} We hypothesize that the backtracking model can think one step ahead without sacrificing its ability to search. Therefore, we shorten the search trace by skipping the last search step. 
        \textit{(b).} Two data variations that improve the backtracking model. Mixed-backtrack model trained on a diverse set of search strategies. Think-backtracking model trained on shortened DFS trace. 
        }
        \label{fig:backtrack_algo}
\end{figure}

\subsection{Two ways to improve backtracking model}
\label{sec:two_variations}
\paragraph{Training on diverse set of search strategies.}
Our analysis suggests a clear direction for improving the backtracking model: using better search strategies to improve backtracking traces. Beyond DFS, we explored alternatives including Breadth-First Search (BFS) and various heuristic methods (see Appendix \ref{appdx:cd_strat}). Despite these efforts, no single search strategy significantly outperformed DFS.
Inspired by \cite{Gandhi2024-tu}, we trained a variant of the backtracking model---\textit{mix-backtrack} model----using a diverse mixture of BFS and DFS strategies (32 in total), aiming to help the model discover more optimal search patterns.

\paragraph{Backtracking model \textit{thinks less and talks more}.}
Apart from learning suboptimal search strategies, another inefficiency in the backtracking model is caused by the model learns to be excessively verbose. Specifically, by requiring the model to explicitly output every step of the DFS, we may prevent it from internalizing part of the reasoning process. Concretely, we hypothesize that for CountDown, the model can internally plan at least one step ahead, allowing it to shorten its explicit reasoning trace without losing its ability to perform DFS. To test  hypothesis, we train a variation---the \textit{think-backtrack} model---on shortened DFS traces, skipping one intermediate step (Fig.~\ref{fig:backtrack_algo}, \textit{A}).

\paragraph{Mix-strategy results.}
Fig.~\ref{fig:backtrack_algo} (\textit{B}) compares this mixed-strategy model against the original backtracking and direct solution models. We also include a training data upper bound, representing perfect execution of the mixed search strategies. The mixed-strategy model improves over the original backtracking model and closely approaches its training-data upper bound. However, even with deliberate attempts to optimize search strategies, surpassing the performance of the direct solution model remains challenging. This experiment underscores the inherent difficulty in identifying superior handcrafted search traces. 

\paragraph{Think-backtrack results.}
Fig.~\ref{fig:backtrack_algo} (\textit{B}) also compares the performance of the think-backtrack model. By encouraging the model to internalize parts of the reasoning process, the think-backtrack model achieves performances comparable to the direct solution model. This result suggests that models with backtracking ability might produce long but ineffective CoT. By training the model to avoid making the mistakes at the first place, we reduce model verbosity without sacrificing its search capability, and in turn improving test-time-compute scaling. As an additional evidence, in Appendix \ref{appdx:confusion}, we show that the think-backtrack model solves a superset of test problems solved by the original backtrack model.

%% file: sections/two_factor.tex
\section{Model size and tree depth impact the efficacy of backtracking}
While we’ve shown that backtracking might lead to ineffective test-time scaling, other factors also shape its effectiveness. In Section~\ref{sec:model_size}, we show that backtracking and direct solution models scale differently with model sizes. To explain the contrasting outcomes (Fig.~\ref{fig:main_figure} \textit{A}) between CountDown and Sudoku, in Appendix~\ref{appdx:game_type}, we show that task differences—particularly search tree depth—play a key role: deeper tasks like Sudoku benefit more from backtracking.

\begin{figure}[t]
    \centering
    \includegraphics[width=1.0\linewidth]{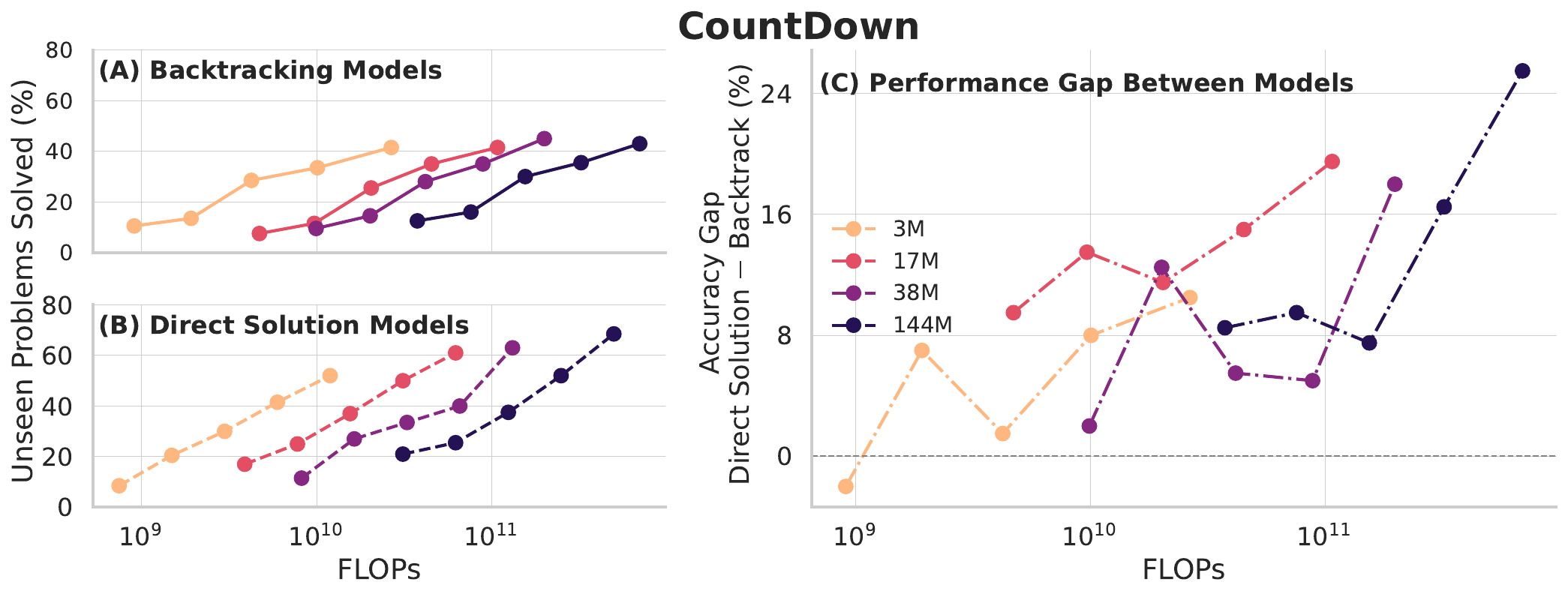}
    \includegraphics[width=1.0\linewidth]{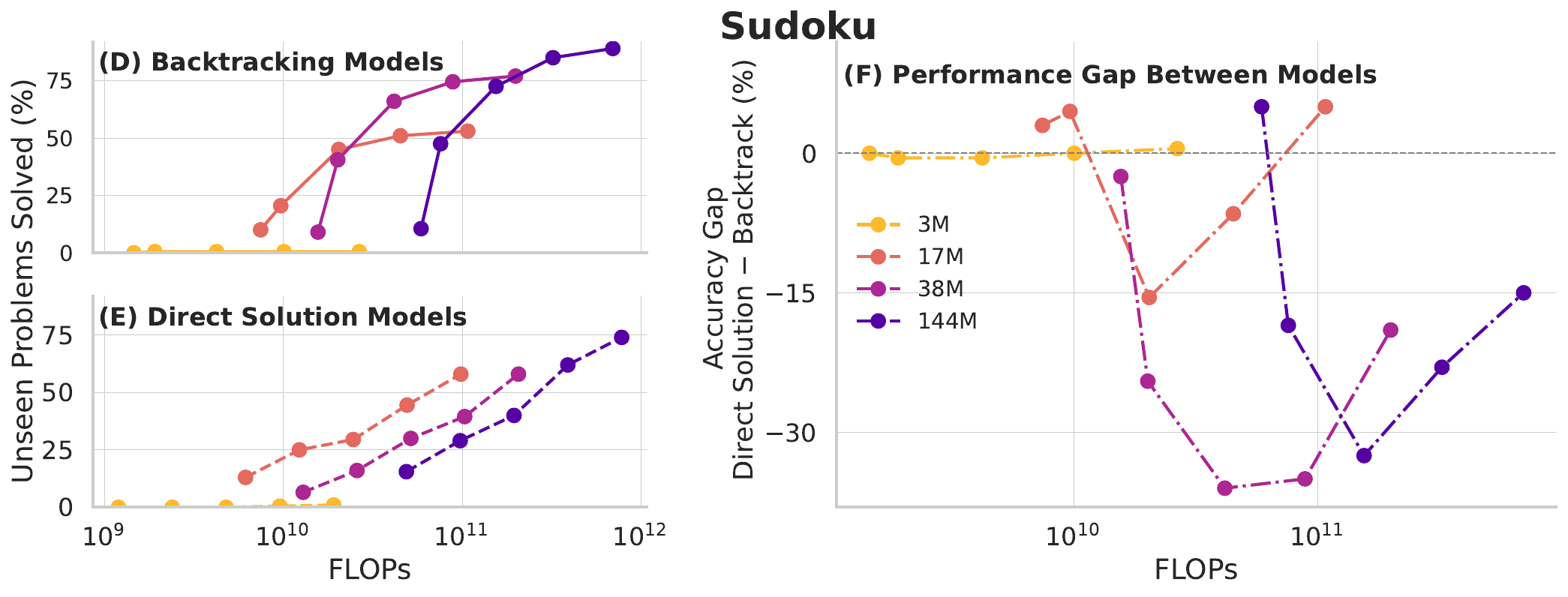}
    \caption{
    \textbf{Different scaling behaviors for backtracking versus direct solution model. }
    \textbf{CountDown} \textit{(A)}. Backtracking model performance does not improve as we scale up model size.
    \textit{(B)}. The direct solution model improves 
    \textit{(C)}. Direct solution model consistently \textit{outperforms} backtracking model. 
    \textbf{Sudoku} \textit{(D, E)}. Both models' performances improve as we scale up model size. 
    \textit{(F)}. Direct solution model consistently \textit{underperforms} backtracking model. 
    }
    \label{fig:model_size}
\end{figure}

\subsection{Dependence on model size}
\label{sec:model_size}
We now investigate how model size impacts the performance of backtracking and direct solution models. We evaluate four model scales—3M, 17M, 38M, and 144M—by proportionally increasing the number of attention heads, embedding dimensions, and number of attention layers. Detailed model configurations can be found in Appendix \ref{appdx:model_scale}.

\paragraph{CountDown.} Scaling up model size improves the performance of the direct solution model (Fig.~\ref{fig:model_size} \textit{B}) across all test-time-compute budgets. When trained exclusively on correct solutions, larger models can independently discover highly effective search strategies. In contrast, the backtracking model shows \textit{no} improvements with increased model sizes (Fig.~\ref{fig:model_size} \textit{A}). The lack of improvement from model scaling can be explained by training data: The performance of backtracking model is constrained by the quality of the backtracking traces used for training. As previously seen in Fig.~\ref{fig:backtrack_algo} (\textit{right}), the 17M backtracking model is already approaching the performance ceiling that is set by the training data. Training larger models on the same backtracking data would not lead to further performance improvements. Due to different scaling behaviors between backtracking and direct solution models, the gap in performances between two types of models widens with increasing model sizes (Fig.~\ref{fig:model_size} \textit{C}).

\paragraph{Sudoku.}
Similar to CountDown, the performances of direct solution models improve with increased model sizes (Fig.~\ref{fig:model_size} \textit{E}). Unlike CountDown, however, the backtracking model also significantly benefits from scaling (Fig.~\ref{fig:model_size} \textit{D}). This difference can again be explained by examining the backtracking training data. Sudoku is inherently more complex than CountDown. The DFS backtracking traces successfully solve 97\% of test boards---far exceeding the current performance of all four tested model sizes. Because the backtracking model for Sudoku has not yet reached training data performance ceiling, increased model capacity leads to improved results. On the other hand, due to the complexity and large search space of the game, the backtracking models' performance gains start to diminish as the search traces become longer. As a result, the backtracking model consistently outperforms the direct solution model across scales, but the advantages diminishes at larger compute budgets (Fig.~\ref{fig:model_size} \textit{E}). \looseness=-1

%% file: sections/rl.tex
\section{GRPO: Learning beyond the imitation game}
\label{sec:rl}
So far, we have shown that under supervised learning, backtracking is not always optimal for scaling test-time compute. We now explore how further training both backtracking and direct solution models with reinforcement learning leads to qualitatively different outcomes.

\subsection{Continue training models with GRPO}
Recently, RL has become a popular approach to further enhance LLMs performance on challenging benchmarks such as MATH \citep{Hendrycks2021-nv} and AIME \citep{AIME2024-ew}. Here, we study the effects of RL in a controlled setting, focusing on how it impacts a model’s backtracking behaviors (sequential search) and as well as a model's parallel search capability (sampling with best-of-$n$). We take the CountDown backtracking and direct solution models, which have been trained to convergence under the supervised learning objective (see Appendix \ref{appdx:training_curve} for training curves). We then continue training each model using GRPO \citep{Shao2024-me}, following verl’s \citep{Sheng2024-bi} implementation. We perform GRPO on the same training data used for the supervised learning. As before, we evaluate performance across different test-time compute budgets.

\subsection{Backtracking model discovers new search strategies}
Figure \ref{fig:main_figure} \textit{C} shows that the backtracking model post GRPO sees an performance boost across all test-compute budgets. The post-GRPO model (\textit{dark red}) reaches an accuracy comparable to the pre-GRPO direct solution model (\textit{light blue}). This improvement is surprising for two reasons: (1) at maximum compute (4096 tokens), the model solves nearly 70\% of the test set—exceeding the performance of the DFS strategy used to generate training data (57\%); and (2) the model was trained on questions it has already seen during supervised learning, with no new problems introduced during GRPO.

\begin{figure}
    \centering
    \includegraphics[height=1.8in]{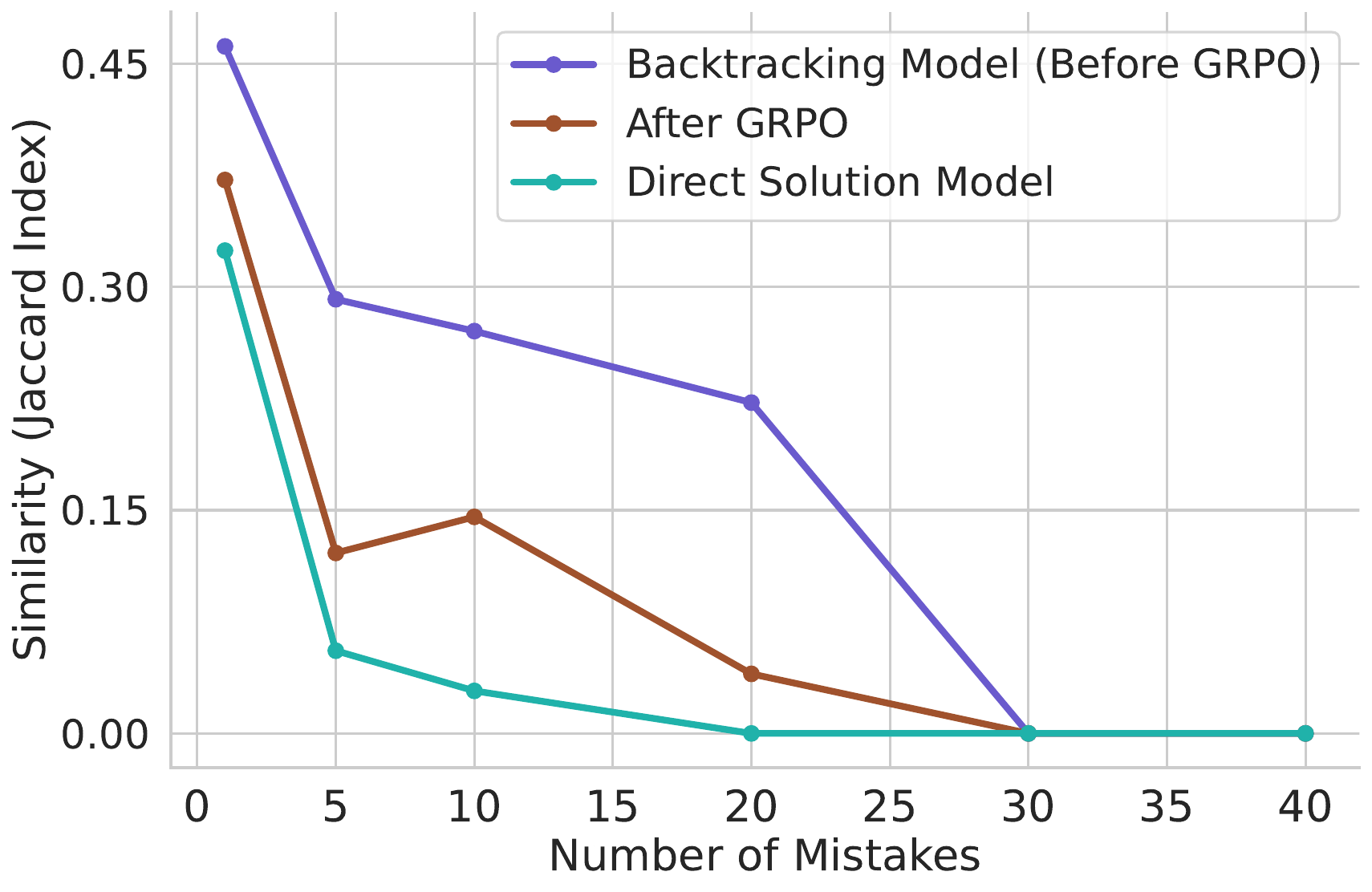}
    \hspace{20px}
    \includegraphics[height=1.8in]{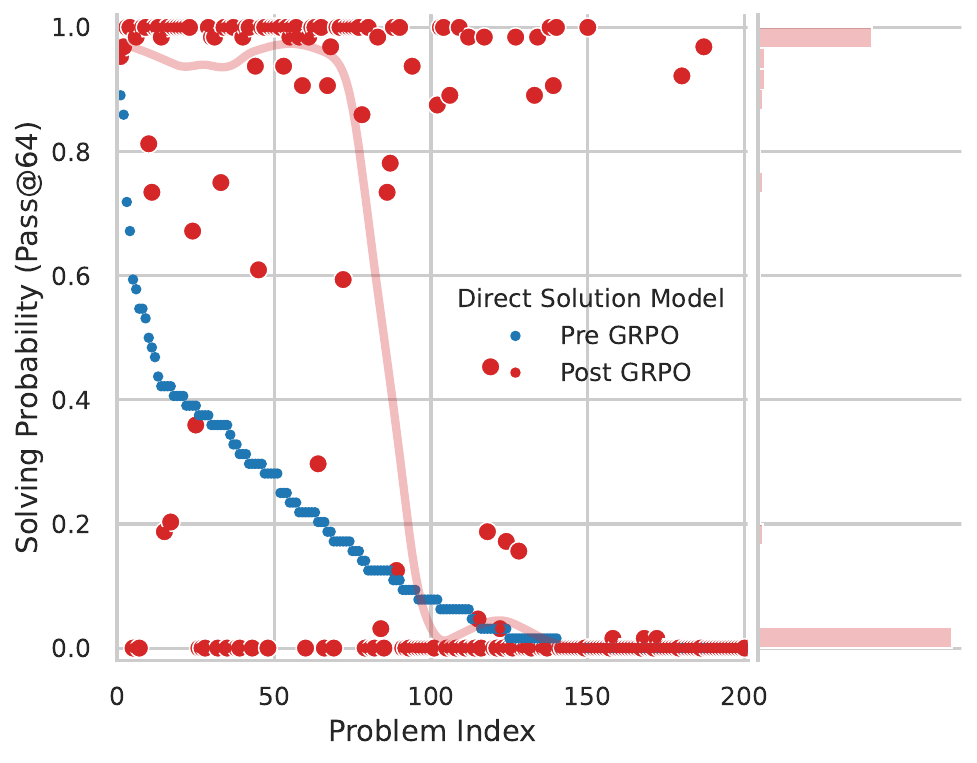}
    \caption{
    \textbf{GRPO has different effect on backtracking versus direct solution model} 
    \textit{Left:} After GRPO, the backtracking model's search strategy starts to deviate away from the DFS search. 
    \textit{Right:} For problems the pre-GRPO direct solution model (\textit{blue}) have a non-zero pass@$k$ solving probabilities, the post-GRPO direct solution model (\textit{red}) solves with pass@1. 
    }
    \label{fig:grpo}
\end{figure}

These gains suggest that the backtracking model, once freed from the constraints of predicting next token on DFS traces, can now discover better search strategies. To concretely show that the backtracking model post-GRPO learns search strategies different from DFS training traces, we revisit the mistake-counting analysis from Section \ref{sec:search_strat} (Figure \ref{fig:countdown_performance_correlation}). For each test problem, we compute the number of mistakes as before (i.e., counting how many incorrect terminal nodes are explored before reaching a correct solution). Using the same set similarity measure as before, we quantify the strategy deviation in Figure \ref{fig:grpo} (\textit{left}). The smaller Jaccard index values confirm that through GRPO, the backtracking model has learned new and more effective search behaviors.
In Appendix \ref{appdx:rl}, we also show the per-problem scatter plot as done in Figure \ref{fig:countdown_performance_correlation}. 

\subsection{Direct solution model specializes at pass@1}
We now show that compared to backtracking models, GRPO has remarkably different effects on direct solution models. As shown in Figure \ref{fig:main_figure} \textit{C}, the direct solution model post-GRPO achieves strong performance at the smallest compute budget (pass@1), solving 42.5\% of unseen CountDown puzzles (82 out of 200 test problems). None of the handcrafted search strategies (Appendix \ref{appdx:cd_strat}) can reach such high accuracy. To understand the impressive gain on 1-shot performance, we examine those 82 problems, and discover that the pre-GRPO direct solution model was able to find correct solution by sampling best-of-$n$ (with $n \leq 64$). We now examine a model's solving probabilities (i.e., measuring pass@$k$ rate out of the 64 generations). We compare the pass@$k$ rate for the diret solution model pre and post GRPO, shown in Figure \ref{fig:grpo}, \textit{right}. We rank the 200 test problems by the pre-GRPO model's solving probabilities. For problems that the pre-GRPO model has a non-zero pass@k rate, the post-GRPO model can solve most of them with pass@1.

However, this improvement in 1-shot performance comes with a substantial trade-off: the model loses its ability to generate diverse solutions. As a result, when we perform parallel search using best-of-$n$, the direct solution model post-GRPO fail to explore different solution paths, hurting its test-time-scaling effectiveness. Therefore, test-time compute scaling becomes ineffective as we increase compute bugdets, forming a sharp contrast to the backtracking model’s consistent improvements across the full compute budget. \looseness=-1

%% file: sections/conclusion.tex
\section{Conclusion and discussions}
In this work, we conducted a controlled empirical investigation into the efficacy of teaching backtracking to large language models (LLMs) as a method for scaling test-time computation. Using two strategic games, CountDown and Sudoku, we demonstrated that backtracking does not universally outperform parallel solution strategies; rather, its effectiveness depends significantly on task characteristics, model scale, and training approach. Appendix~\ref{appdx:real_math}, we show that our resuls in synthetic setting generalize: even in real-world reasoning tasks, backtracking is \textit{not} always beneficial. Additionally, our reinforcement learning experiments uncovered a unique synergy between backtracking capabilities and RL-based training, enabling models to discover novel strategies.

\paragraph{Limitations and future work.}
While our experiments relied on two strategic games (CountDown and Sudoku) and models trained from scratch—common practices for controlled studies—an important avenue for future research is extending our findings to complex, real-world reasoning tasks such as coding and mathematical problem-solving. 
For future work, developing precise ways to characterize tasks that benefit from backtracking will be valuable for guiding model training. Finally, while we intentionally created a dichotomy between pure backtracking and direct-solution models, real-world applications may require hybrid strategies that dynamically choose between direct generation and explicit search based on problem complexity. Investigating whether LLMs can autonomously optimize their reasoning modes, particularly through reinforcement learning paradigms, is a promising future direction for improving the flexibility and efficiency of model reasoning.


%% file: sections/appendix/real_math.tex
\section{Backtracking Analysis on Math Reasoning with LLMs}
\label{appdx:real_math}
\begin{figure}[t!]
    \centering
    \includegraphics[height=0.35\linewidth]{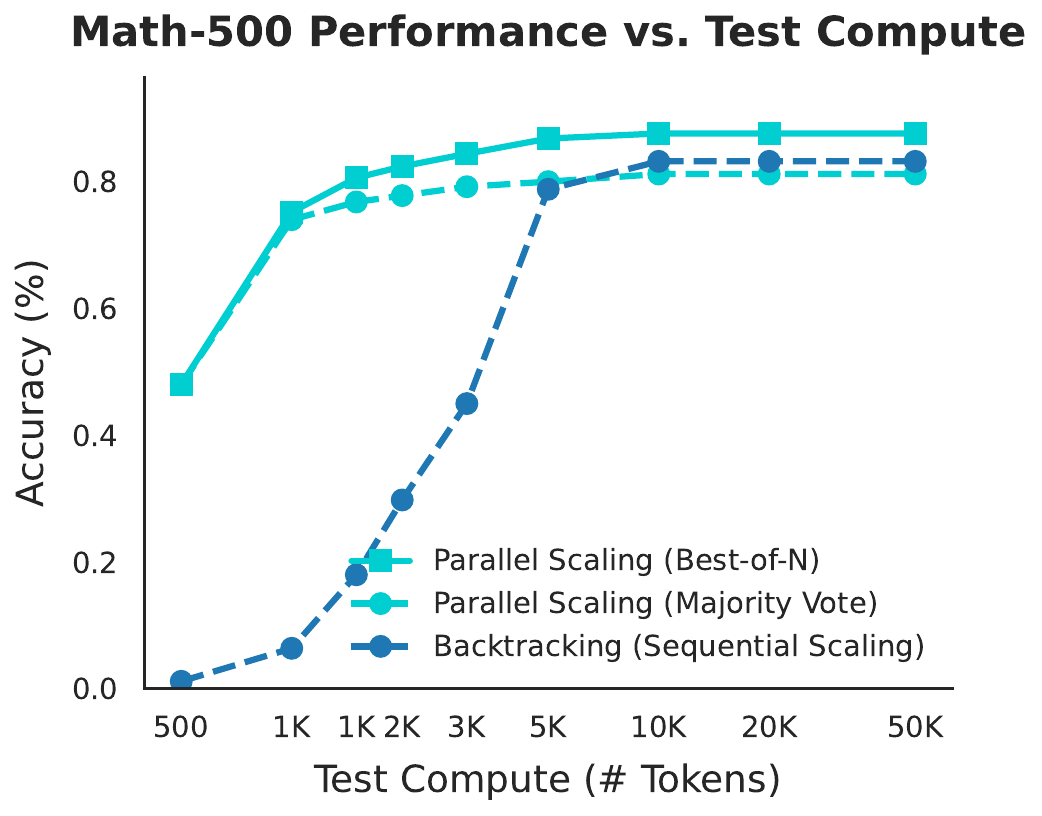}
    \includegraphics[height=0.35\linewidth]{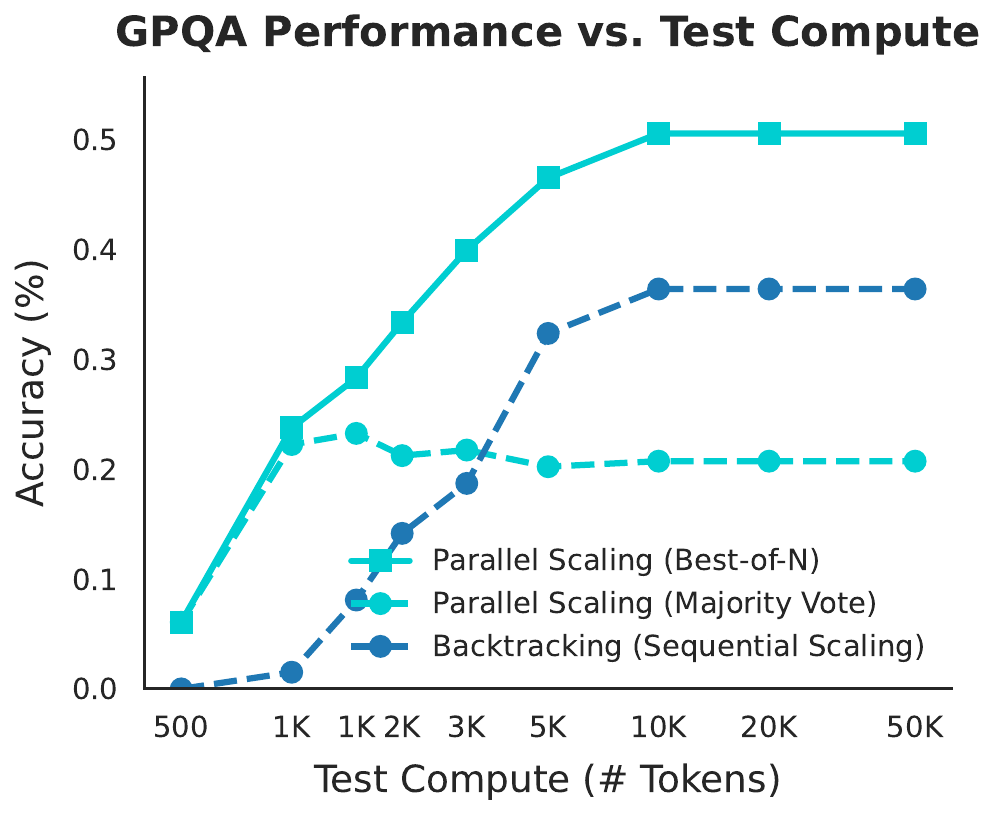}
    \caption{
    \textbf{Evaluating backtracking on real LLMs.} 
    \textit{Left:} On MATH-500, we compare the S1 model (fine-tuned on backtracking traces) using sequential decoding with budget forcing, against its base model (Qwen2.5-32B-Instruct) using parallel sampling. The backtracking model \textit{underperforms} at low compute but narrows the gap at higher budgets. 
    \textit{Right:} On GPQA, the same backtracking setup \textit{outperforms} parallel sampling in a multiple-choice reasoning setting. 
    \textbf{This comparison generalizes our conclusion from synthetic settings to real LLMs. }
}
    \label{fig:real_math}
\end{figure}
\subsection{Experimental Setup}

To complement our synthetic experiments, we conduct an evaluation on real-world math problems to examine whether \textbf{backtracking remains effective under equal test-time compute}. We compare two approaches:

\begin{itemize}[itemsep=2pt,labelindent=2pt,topsep=0pt,parsep=0pt,partopsep=1pt, align=left, leftmargin=*]
    \item \textbf{Backtracking model}: fine-tuned on solution traces that include explicit self-correction and step-by-step reflection.
    \item \textbf{Direct solution model}: the base model without backtracking fine-tuning, using parallel sampling (with majority voting for final correct answer) at inference.
\end{itemize}

To control test-time compute, we use the \textbf{budget forcing} technique introduced in \citep{Muennighoff2025-ms}. This enables a fair comparison across models with differing reasoning styles.

\paragraph{Backtracking Model.} We adopt the \texttt{S1} checkpoint from \cite{Muennighoff2025-ms}, a model trained on solution traces distilled from DeepSeekR1. These traces exhibit explicit backtracking behaviors---identifying and correcting earlier mistakes. We generate outputs with temperature $T = 0.7$ under budget forcing and evaluate on the \texttt{MATH-500}.

\paragraph{Direct Solution Model.} For fair comparison, we use the same base model as \texttt{S1}---Qwen2.5-32B-Instruct---without backtracking fine-tuning. We sample $N=1$ to $8$ completions with temperature $T = 0.7$, and report both \textbf{Best-of-$N$} and \textbf{Majority Vote} accuracy.

\subsection*{Results and Interpretation}

Figure~\ref{fig:real_math} (\textit{left}) presents accuracy under matched compute budgets. We observe that at \textbf{low compute budgets} the backtracking model underperforms due to its verbose reasoning traces. At \textbf{higher budgets}, backtracking matches and slightly exceeds the performance of parallel sampling.
This mirrors trends observed in the \textbf{CountDown} (Section~4.3), and suggests that while backtracking introduces overhead, it yields benefits when sufficient compute is available.

To form a sharp contrast, we reproduce results from \citep{Muennighoff2025-ms} on \texttt{GPQA-Diamond} (Figure~\ref{fig:real_math}, \textit{right}), which shows that the same backtracking model significantly outperforms parallel sampling---even at lower budgets---in a multiple-choice setting. This contrast highlights that \textbf{the effectiveness of backtracking is task-dependent}.

This real-world evaluation supports our synthetic findings: \textbf{backtracking improves performance under compute constraints}, but its advantage depends on the task structure. On open-ended math problems, the benefit is most pronounced at higher budgets. On structured tasks like multiple-choice QA, gains can appear even earlier. Overall, our conclusions \textbf{generalize beyond synthetic settings}.

%% file: sections/appendix/related.tex
\section{Related Work Extended}
\label{appdx:related}
\subsection{Test-time computation scaling}
A growing body of work has explored how to improve language model performance by scaling test-time computation. These approaches typically fall into two broad categories: \textbf{parallel} and \textbf{sequential} search. Parallel methods sample multiple solutions independently and select the best one using predefined criteria—such as majority voting or external reward models—as seen in Best-of-$N$ techniques \citep{Brown2024-zj, Irvine2023-jv, Levi2024-wd}. These methods often rely on outcome-based reward models that score complete solutions \citep{Xin2024-rj,  Ankner2024-mc}.

In contrast, sequential methods iteratively refine reasoning by conditioning on previous attempts. This class includes stepwise improvement methods\citep{Ankner2024-mc, Hou2025-qz, Lee2025-tw}, where each new trajectory builds on earlier outputs, enabling the model to adapt its reasoning dynamically. Other research works have also explored using the search process itself to improve model reasoning capabilities, either during inference or by integrating the feedback into training\citep{Wang2024-up, Luo2024-on}. While these methods can reduce redundancy, they typically require more compute per sample and may suffer from compounding errors.

Tree-based approaches, such as Monte Carlo Tree Search (MCTS) and guided beam search, represent a hybrid between parallel and sequential strategies\citep{Gandhi2024-tu, Liu2023-os, Zhang2023-ai, Zhou2023-dz, Choi2023-ji, Xie2023-vx}. These methods often leverage process reward models, which assign value to intermediate reasoning steps rather than full outputs\citep{Lightman2023-cg, Wang2024-up, Wu2024-bj}. REBASE\citep{Wu2024-bj}, for example, uses a process reward model to guide exploration and pruning in tree search, and has been shown to outperform both sampling-based methods and traditional MCTS.

\subsection{Self-correction and backtracking}
Search and backtracking are inherently tied to self-correction, as they enable models to revisit earlier decisions and recover from errors—a critical capability for multi-step reasoning. Teaching language models to self-correct has been approached through fine-tuning on revision demonstrations from humans or stronger models\citep{Saunders2022-cr, Ye2023-zx, Qu2024-rl}, as well as through synthetic data generation and handcrafted augmentation\citep{Paul2023-sj, Wang2023-pd, Lee2023-gv}. Reward-based methods provide another avenue, using outcome- or process-level signals to differentiate good and bad reasoning trajectories, often framed as implicit policy learning\citep{Welleck2022-jb, Akyurek2023-dp, Zhang2024-iz}. Some methods further incorporate search, critique generation, or separate correction modules to enhance reasoning quality\citep{Yao2023-qu, Havrilla2024-bl}. In contrast, using two structured games, we investigate the tradeoffs of teaching models to backtrack via search traces versus allowing them to learn purely from correct solutions.

\subsection{Reinforcement learning for LLM reasoning}
Reinforcement learning (RL) has emerged as a powerful framework for improving the reasoning abilities of language models. While early work applied off-policy and on-policy RL methods to guide models toward verifiable outcomes\citep{Zelikman2022-yo, Kazemnejad2024-gw}, recent approaches have shown that even simplified algorithms like GRPO can lead to significant performance gains and the emergence of in-context search behavior \citep{DeepSeek-AI2025-vs, Shao2024-me, DeepSeek-AI2025-vs}. These advances suggest that RL can help models autonomously discover more effective reasoning strategies, even without explicit reward models or structured search. However, not all models benefit equally from RL, and it remains unclear what properties make a model amenable to learning through reinforcement. Our work contributes to this question by examining how backtracking models, when trained with GRPO, can discover novel solution strategies—while no-backtracking models show limited or mixed gains.

%% file: sections/appendix/hyperparameter.tex
\section{Experiment details}
\label{appx:set_up}
\subsection{Additional details on game, data generation}
\label{appdx:data_detail}

\paragraph{CountDown tree size computation.} CountDown has an exponentially growing search space with respect to the number of candidate numbers. If the current state has $N$ available numbers, there are $\binom{N}{2} \times 4$ possible actions (selecting a pair and one of four operations), and the depth of the tree is $N - 1$. For games with four candidate numbers, the complete search tree contains 1,152 nodes.

\paragraph{CountDown search.}
To generate DFS search data, we use a sum heuristic to guide the search order and prune nodes. This heuristic measures the distance between the sum of all input numbers and the target number, and prunes a node if the absolute distance exceeds the target. This approach is inspired by \cite{Gandhi2024-tu}, who also consider an alternative—the multiply heuristic—which measures the minimum distance between the input set and the factors of the target. However, in our experiments, both heuristics yield similar performance: for a fixed number of visited nodes, DFS with either heuristic solves approximately the same number of games.

\paragraph{Sudoku rule.} In a Sudoku game, the player is given a $9 \times 9$ grid in which each cell must be filled with a digit from 1 to 9. The puzzle is subject to three constraints: each row, each column, and each of the nine $3 \times 3$ subgrids must contain all digits from 1 to 9 exactly once. Given a partially filled grid, the objective is to fill in the remaining cells such that all constraints are satisfied. 

\paragraph{Sudoku data and tokenization.} To represent the Sudoku board for language models, we encode each cell as a position-value pair: $(x, y) = v$, where $(x, y)$ denotes the grid location and $v$ is the cell’s value. The model receives the initial board as a list of known $(x, y) = v$ pairs and generates the solution by predicting the values for the remaining cells. We generate \textbf{backtracking traces} by serializing the full DFS traversal. For the \textbf{direct solution model}, we prune each trace to include only the final solution path. 

\paragraph{Scoring.} For CountDown, a solution is correct only if it adheres to game rules and correctly achieves the target number. For Sudoku, correctness requires fully solving the board, with no partial credit given for incomplete but correct boards. Models are tested on 200 unseen problems per game. The same scoring function is used as the reward function in GRPO (Section \ref{sec:rl})

\subsection{Additional details on model architecture}
\label{appdx:model_scale}
Model hyperparameters can be found in Table \ref{tab:model_sizes}.

\begin{table}[ht]
    \centering
    \begin{tabular}{@{}lccccc@{}}
    \toprule
    \textbf{Model Size} & \textbf{Hidden Size} & \textbf{Layers} & \textbf{Attn Heads} & \textbf{Intermediate Size} & \textbf{KV Heads} \\ \midrule
    3M   & 256  & 6  & 4 & 512  & 1 \\
    17M  & 512  & 8  & 4 & 1024 & 1 \\
    38M  & 512  & 10 & 8 & 2048 & 2 \\
    144M & 1024 & 12 & 8 & 3072 & 2 \\ \bottomrule
    \end{tabular}
    \caption{Qwen2.5-style architecture configurations for the four model sizes used in our experiments.}
    \label{tab:model_sizes}
\end{table}

\subsection{Training hyperparameter}
\label{appdx:hparam_train}
Training hyperparameters can be found in Table \ref{tab:training_hyperparams}. We train all models on 2 NVIDIA H100 80GB HBM3 GPUs.

\begin{table}[ht]
    \centering
    \begin{tabular}{@{}ll@{}}
    \toprule
    \textbf{Hyperparameter} & \textbf{Value} \\ \midrule
    \multicolumn{2}{l}{\textit{Optimization}} \\
    Learning rate & $1 \times 10^{-5}$ \\
    Weight decay & 0.01 \\
    
    \multicolumn{2}{l}{\textit{Learning Rate Schedule}} \\
    Scheduler type & Cosine \\
    Warmup steps & 1 \\
    
    \multicolumn{2}{l}{\textit{Training Setup}} \\
    Epochs & 30 \\
    Batch size (backtracking model) & 32 \\
    Batch size (direct solution model) & 64 \\
    Context length (backtracking model) & 4096 \\
    Context length (direct solution model) & 512 \\
    
    \multicolumn{2}{l}{\textit{Tokenizer}} \\
    Tokenizer size (CountDown) & 74 \\
    Tokenizer size (Sudoku) & 110 \\ \bottomrule
    \end{tabular}
    \caption{Training hyperparameters used for all experiments. Batch size and context length vary based on model type.}
    \label{tab:training_hyperparams}
\end{table}

%% file: sections/appendix/compute_flops.tex
\section{FLOP computation}
\label{appdx:compute_flops}
To compare backtracking and direct solution models under a fixed compute budget, we estimate inference FLOPs based on model architecture and generation length $T$. We use a simplified transformer FLOP computation that accounts for per-token operations across all layers.

Below is a list of architectural and generation parameters:
\begin{itemize}[itemsep=2pt,labelindent=2pt,topsep=0pt,parsep=0pt,partopsep=1pt, align=left, leftmargin=*]
    \item $d_{\text{model}}$: hidden dimension
    \item $d_{\text{kv}}$: key/value dimension \footnote{key/value dimension is different from hidden dimension because of GQA \citep{Ainslie2023-wh}}
    \item $d_{\text{ff}}$: intermediate (feedforward) dimension
    \item $L$: number of layers
    \item $T$: number of generated tokens (i.e., context length)
    \item $N$: number of sequences generated (e.g., in best-of-$N$ sampling)
\end{itemize}

\subsection{Step-by-step FLOPs Calculation}

\paragraph{1. Per-layer linear FLOPs per token.}
We break down the linear FLOPs for each transformer layer into attention and MLP components:

\begin{itemize}[itemsep=2pt,labelindent=2pt,topsep=0pt,parsep=0pt,partopsep=1pt, align=left, leftmargin=*]
    \item \textbf{Self-attention:}  
    \begin{itemize}
        \item Query projection: $d_{\text{model}} \times d_{\text{model}}$
        \item Key projection: $d_{\text{model}} \times d_{\text{kv}}$
        \item Value projection: $d_{\text{model}} \times d_{\text{kv}}$
        \item Output projection: $d_{\text{model}} \times d_{\text{model}}$
    \end{itemize}
    This results in a total of:
    \[
    \text{FLOPs}_{\text{attention-linear}} = 2 d_{\text{model}}^2 + 2 d_{\text{model}} d_{\text{kv}}
    \]

    \item \textbf{MLP (Feedforward):}  \\
    MLP layers include following components:
    \begin{itemize}[itemsep=2pt,labelindent=2pt,topsep=0pt,parsep=0pt,partopsep=1pt, align=left, leftmargin=*]
        \item Gate projection
        \item Up projection
        \item Down projection
    \end{itemize}
    Each of these MLP layers costs: $d_{\text{model}} \times d_{\text{ff}}$, giving:
    \[
    \text{FLOPs}_{\text{mlp}} = 3 d_{\text{model}} d_{\text{ff}}
    \]
\end{itemize}

Combining both components, the total per-token linear cost per layer is:
\[
\text{FLOPs}_{\text{linear}} = 2 d_{\text{model}}^2 + 2 d_{\text{model}} d_{\text{kv}} + 3 d_{\text{model}} d_{\text{ff}}
\]

\paragraph{2. Quadratic attention cost.} 
Self-attention involves computing interactions between all token pairs, resulting in a quadratic cost:
\[
\text{FLOPs}_{\text{attention}} = d_{\text{model}} \cdot \frac{T(T+1)}{2}
\]

\paragraph{3. Total generation cost per sequence.}
Each token attends to all previous tokens across all $L$ layers. The generation cost for a single sequence is:
\[
\text{FLOPs}_{\text{gen}} = L \cdot \left( \text{FLOPs}_{\text{linear}} \cdot T + \text{FLOPs}_{\text{attention}} \right)
\]

\paragraph{4. Total inference FLOPs.}
For $N$ sequences (e.g., best-of-$N$ sampling), the total inference cost is:
\[
\text{FLOPs}_{\text{total}} = N \cdot \text{FLOPs}_{\text{gen}}
\]

We do not include auxiliary operations such as token embedding and softmax, weight norm, as their contribution is negligible compared to the transformer layers. All FLOPs reported in our experiments use this formula, with model configurations listed in Table~\ref{tab:model_sizes}.

%% file: sections/appendix/majority_vote.tex
\section{Majority voting versus best-of-n}
\label{appdx:majority_vote}
In this work, we primarily use the best-of-$n$ metric to evaluate the direct solution model. This metric is suitable for tasks where verifying the correctness of a solution is trivial, whereas solving the task itself is challenging. Many real-world problems, such as coding tasks and combinatorial optimization, fall into this category. Conversely, for problems where verification is difficult, metrics such as majority voting may be more appropriate.

\begin{figure}[t]
    \centering
        \includegraphics[width=0.5\linewidth]{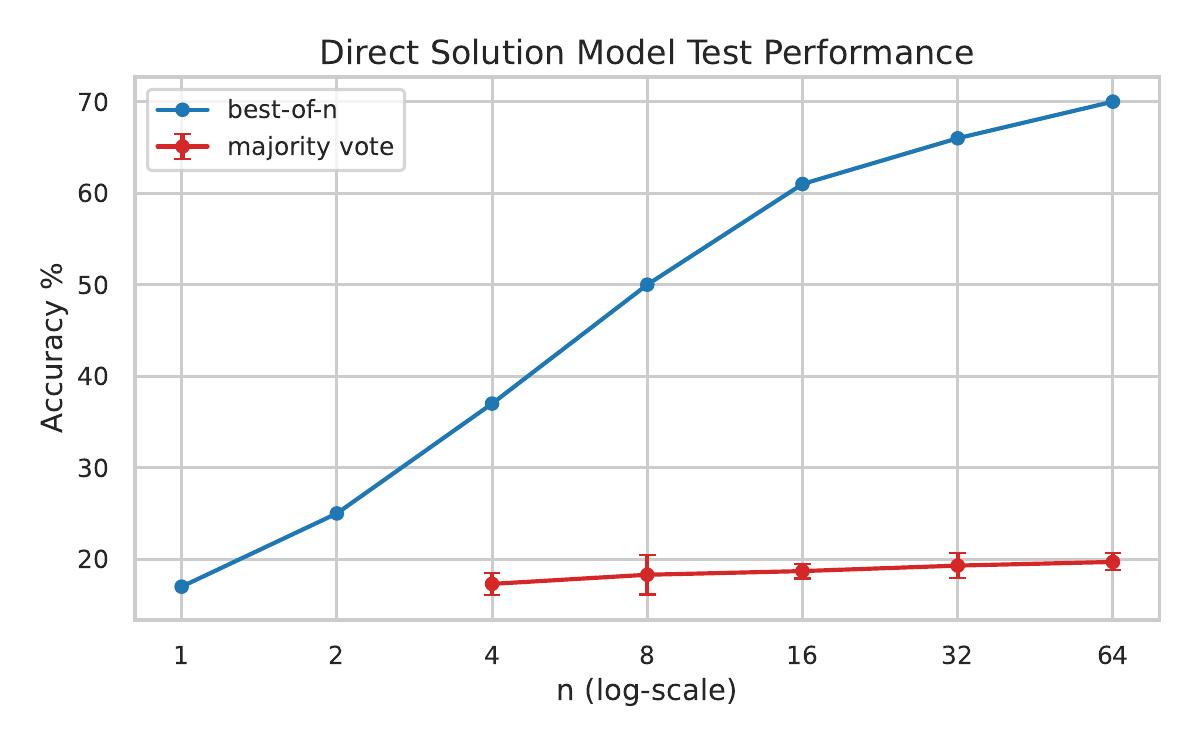}
    \caption{
    \textbf{Majority voting versus best-of-$n$ for CountDown direct solution model.} 
    For CountDown, verification is much easier than solving the problem. Therefore, best-of-$n$ as a performance is justified. Additionally, we also examine majority voting performance. However, CountDown solutions are not unique, majority voting is not the most suitable way to measure model performances. 
    }
    \label{fig:cd_majority_vote}
\end{figure}

To illustrate this point, we additionally evaluate the CountDown direct solution model using both metrics in Figure \ref{fig:cd_majority_vote}. For majority voting, we generate n solutions per test problem, select the most frequently occurring solution (breaking ties randomly), and evaluate its correctness.

We find that the majority-voting performance closely approximates the direct solution model’s one-shot accuracy (i.e., best-of-$n$ with $n$=1). However, majority voting is less suitable for our task for several reasons. First, the CountDown game frequently has multiple correct solutions, so selecting the majority solution path can fail to detect cases where the model generates different but equally valid solutions. Second, while majority voting is appropriate in real-world LLM scenarios—such as mathematical reasoning—where distinct solution paths converge to the same final boxed answer, in our synthetic setting, where models are trained from scratch, majority voting essentially becomes a noisy proxy for greedy decoding (sampling at temperature $T=0$). Thus, we expect and observe majority voting accuracy to closely track pass@1 accuracy.

In summary, given the characteristics of our task and the controlled experimental setup, best-of-$n$ remains a valid and preferred metric for evaluating direct solution models.

%% file: sections/appendix/game_type.tex
\section{Dependence on depth of the search tree}
\label{appdx:game_type}

\begin{figure}
    \centering
    \includegraphics[width=0.4\linewidth]{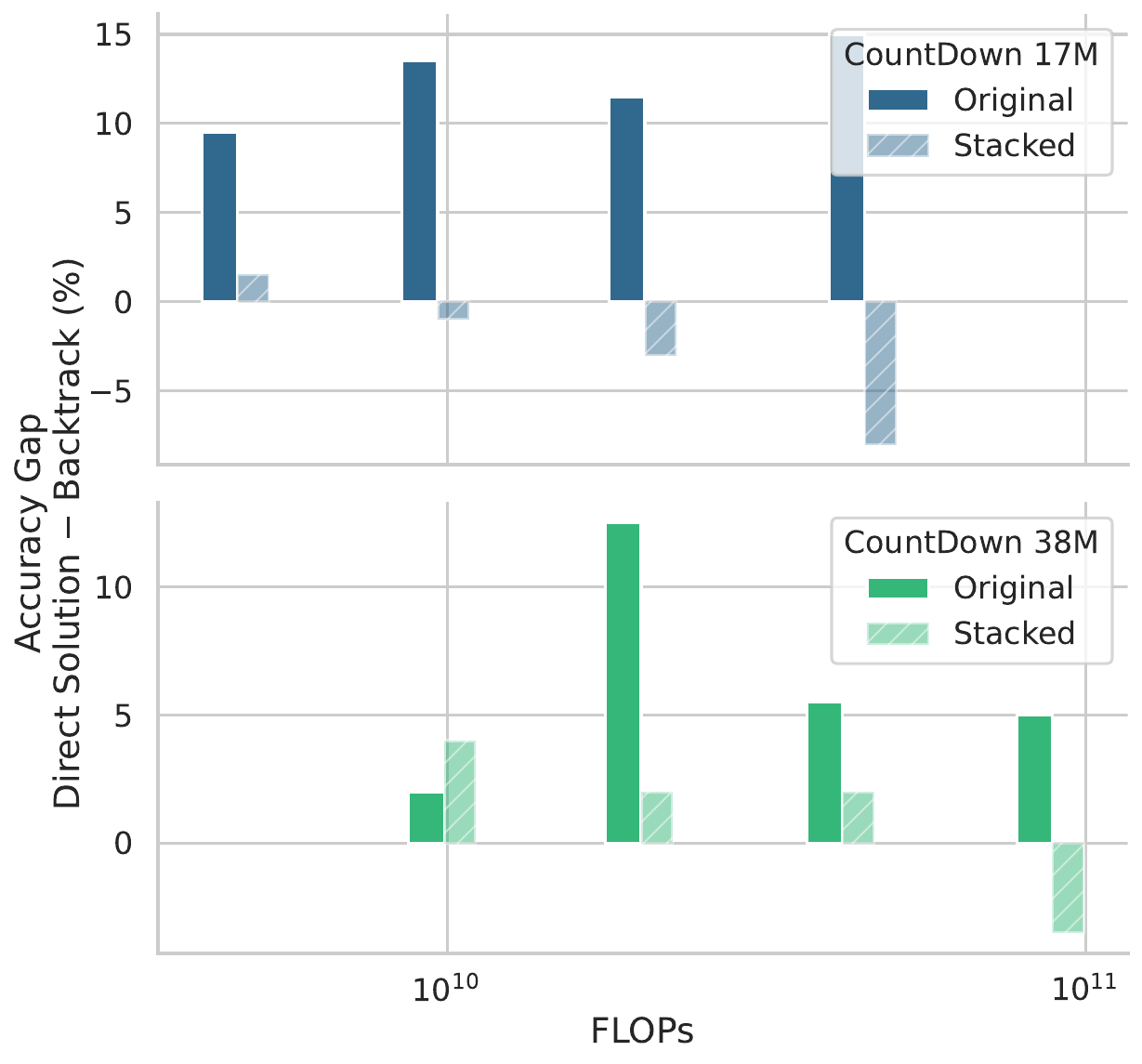}
    \hspace{10px}
    \includegraphics[width=0.4\linewidth]{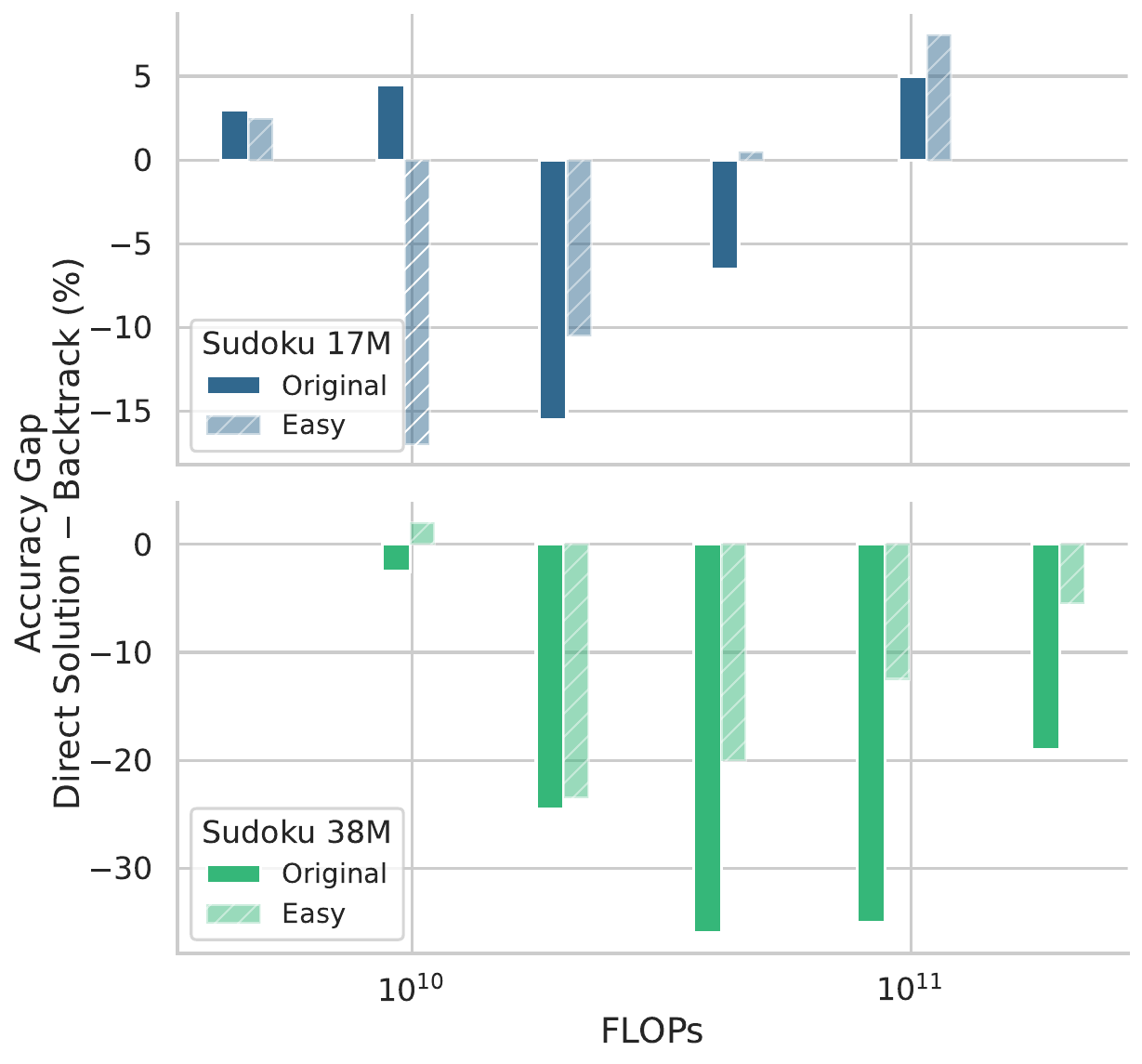}
    \caption{
    \textbf{The efficacy of backtracking depends on search tree depth.} 
    \textit{Left:} We introduce a variation of the CountDown game—stacked CountDown—to \textbf{increase} the search tree depth. In the original CountDown game (\textit{solid bars}), the direct solution model consistently \textbf{outperforms} the backtracking model, shown by a positive performance gap. In the stacked version (\textit{slanted bars}), this gap is significantly reduced or even reversed, indicating that backtracking becomes more compute-efficient at greater depths. 
    \textit{Right:} We introduce a variation of Sudoku—easy Sudoku—where the initial board has more pre-filled cells to \textbf{decrease} the search tree depth. In the original setting, the 38M direct solution model (\textit{bottom, solid bars}) \textbf{underperforms} the backtracking model. In the shallow Sudoku variant (\textit{slanted bars}), the performance gap narrows across compute budgets. For the 17M models (\textit{top}), the results are less conclusive.
}
    \label{fig:tree_depth}
\end{figure}

\subsection{Search tree depth}
Why do backtracking models perform well on Sudoku but underperform on CountDown, even when both are trained on DFS search traces? We argue that task characteristics---particularly those beyond our control in real---world settings—play a key role in determining whether backtracking is test-time-compute-efficient. A major difference between the two games lies in the depth of their search trees (Figure~\ref{fig:setup_demo}). In hard Sudoku puzzles, only 20 out of 81 cells are pre-filled, leaving 50–60 cells to solve. This results in deep search trees with extensive trial-and-error, with many backtracking steps. In contrast, CountDown (in our setup) uses 4 candidate numbers, limiting the search tree depth to just 3. We hypothesize that backtracking models excels at tasks with deeper search trees, while shallow trees make parallel strategies (i.e., direct solution model) more effective. To test this, we design a variant of CountDown with increased search depth and a variant of Sudoku with reduced depth.

\subsection{A deeper CountDown}
\paragraph{Set up} 
To increase the search tree depth in CountDown, one might naively scale up the number of candidate numbers. However, this approach quickly leads to exponential growth in tree width: with 4 candidates, the tree contains 1,152 nodes; with 5 candidates, it grows to 46,080. To prevent the exponential growth in the number of search paths, we design a stacked CountDown variant that increases depth while controlling tree width. In this setup, the player is given 8 candidate numbers and a final target. The first 4 numbers must be used to reach the 5th number ("a partial goal"), and the remaining 4 numbers must then be used to reach the final target. This effectively stacks two CountDown problems, increasing depth without combinatorial explosion. We generate training data for both backtracking and no-backtracking models following the same procedure as in Section~\ref{sec:cd_data}, with examples provided in Appendix~\ref{appx:verbatim} (Figure \ref{fig:cd_stack_data}). We train a 17M as well as a 38M model until validation loss has converged, and test on 200 unseen problems.

\paragraph{Results}
In Figure~\ref{fig:tree_depth} (\textit{left}), we compare the performance gap between the direct solution model and the backtracking model, measured by the difference in test accuracy. In the original CountDown setting (\textit{solid bars}), the direct solution model consistently \textbf{outperforms} the backtracking model across all test compute budgets. However, in the stacked CountDown variant (\textit{slanted bars}), the performance gap narrows significantly---and in some cases, reverses. The sign reverse indicates the backtracking model now \textbf{outperforms} the direct solution model. These results support our hypothesis: in CountDown, backtracking becomes more compute-efficient as the search tree depth increases. We observe this trend across both 17M and 38M models.

\subsection{A shallower Sudoku}
\paragraph{Set up} 
To reduce the search tree depth in Sudoku, we generate easier boards by increasing the number of initially filled cells. Specifically, we take the original 3M Sudoku dataset \cite{Radcliffe2020-as} and apply the direct solution model (Section~\ref{sec:sudoku_data}) to correctly fill 10 additional cells. This increases the average number of pre-filled cells from 20 to around 30, effectively decreasing search tree depth. We generate both backtracking and direct solution training data following the same procedure in Section~\ref{sec:sudoku_data}. Models with 17M and 38M parameters are trained to convergence and evaluated on 200 unseen problems.

\paragraph{Results}
In Figure~\ref{fig:tree_depth} (\textit{right}), we show the performance gap between the direct solution and backtracking models, measured by the difference in test accuracy. In the original (hard) Sudoku setting, the 38M direct solution model consistently \textbf{underperforms} the backtracking model, as indicated by the negative gaps (\textit{solid green bars}). In the shallow-Sudoku variant (\textit{slanted bars}), these gaps are reduced across all test-time compute budgets for the 38M model. The trend is less clear for the 17M model, where the performance difference remains small in both settings. Overall, these results support our hypothesis: in Sudoku, backtracking becomes more test-time-compute-efficient when the search tree is deeper.

%% file: sections/appendix/cd_search_strat.tex
\section{Additional results}
\subsection{Exploring different CountDown strategies}
\label{appdx:cd_strat}
\begin{figure}[t]
    \centering
        \includegraphics[width=0.7\linewidth]{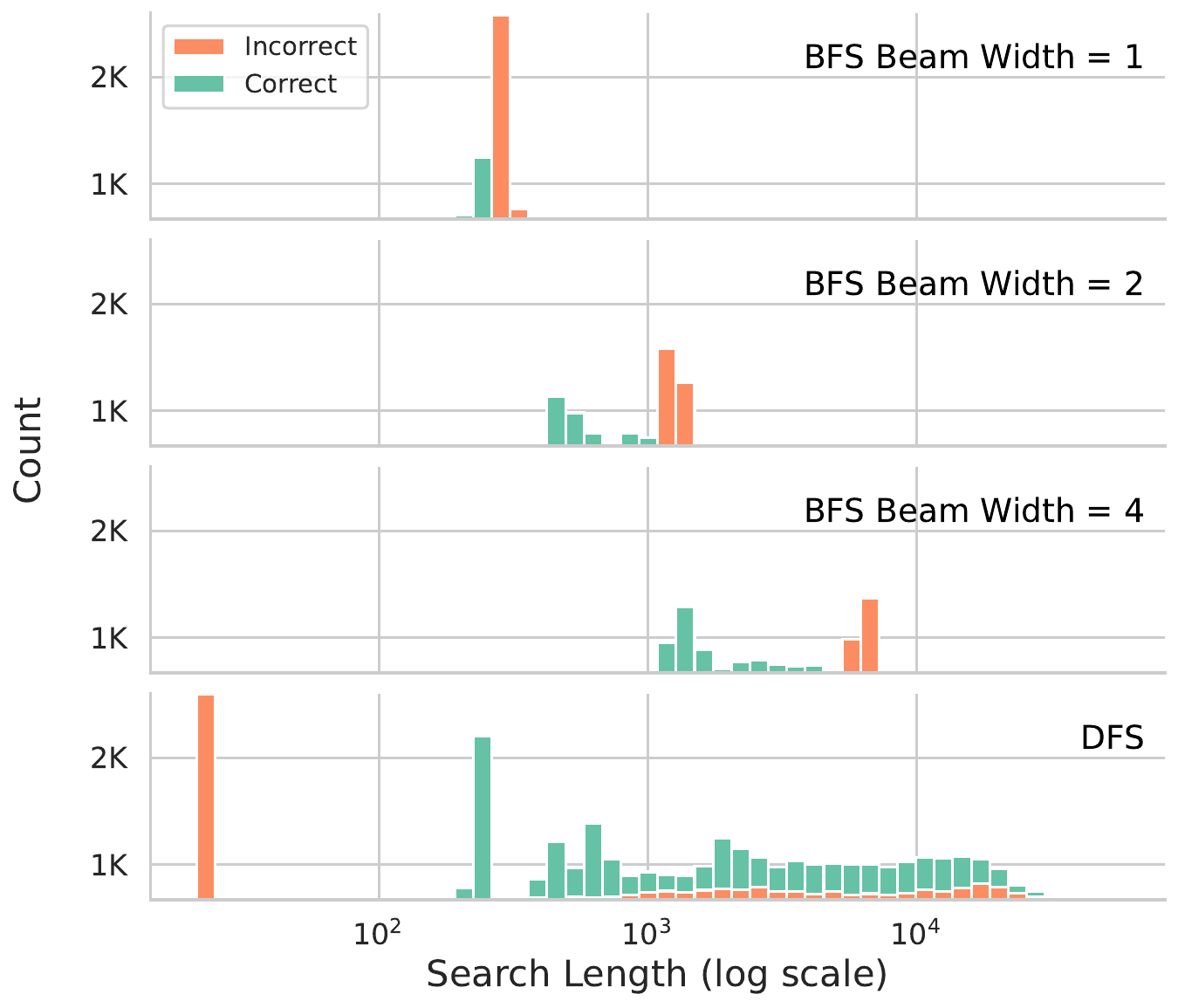}
    \caption{
    \textbf{Examine different search strategies for CountDown.} 
    Beyond DFS, we experiment with Bread-First-Search (BFS) with different beam widths. We tokenize the search trace and measure the number of tokens as search length. 
    There is not one search algorithm that is optimal to generate both short and correct solution traces. 
    }
    \label{fig:cd_search_strat}
\end{figure}

We analyze different search strategies for CountDown, including DFS and BFS with varying beam widths. For each strategy, we tokenize the resulting backtracking trace and measure number of tokens used in each search trace. The  goal is to identify which strategy that finds correct solutions with the fewest tokens (Figure~\ref{fig:cd_search_strat}). The results show no clear winner. BFS with a smaller beam width produces shorter traces by exploring fewer nodes, but this comes at the cost of missing correct solutions more frequently. Increasing the beam width improves solution coverage but leads to longer traces due to broader exploration.

In contrast, DFS produces more uniformly distributed trace lengths but suffers from a specific failure mode: it may prune the correct path early and terminate prematurely. These failures appear as short but incorrect traces, visible as the left-most orange bars in Figure~\ref{fig:cd_search_strat} (\textit{bottom}).

%% file: sections/appendix/confusion_table.tex
\subsection{Compare think-backtrack and backtrack}
\label{appdx:confusion}
Table \ref{tab:confusion} further shows a confusion matrix comparing the original and think-backtrack models. The backtracking model solves 102 test problems in total with maximum test-time compute budget (4096 tokens). Out of those 102 problems, the think-backtrack model solves most of them. This evidence further shows that by training on shortened search traces, the model learns to internalize parts of its thinking without sacrificing performances. 
\begin{table}[h]
    \centering
    \begin{tabular}{lcc}
    \toprule
     & \textbf{T-B Correct} & \textbf{T-B Incorrect} \\
    \midrule
    \textbf{B Correct} & 83 & 19 \\
    \textbf{B Incorrect} & 41 & 57 \\
    \bottomrule
    \end{tabular}
    \caption{Confusion matrix between Think-Backtrack (T-B) and Backtrack (B) models.}
    \label{tab:confusion}
\end{table}

%% file: sections/appendix/training_curve.tex
\subsection{Supervised learning training curve}
\begin{figure}
    \centering
    \includegraphics[width=0.7\linewidth]{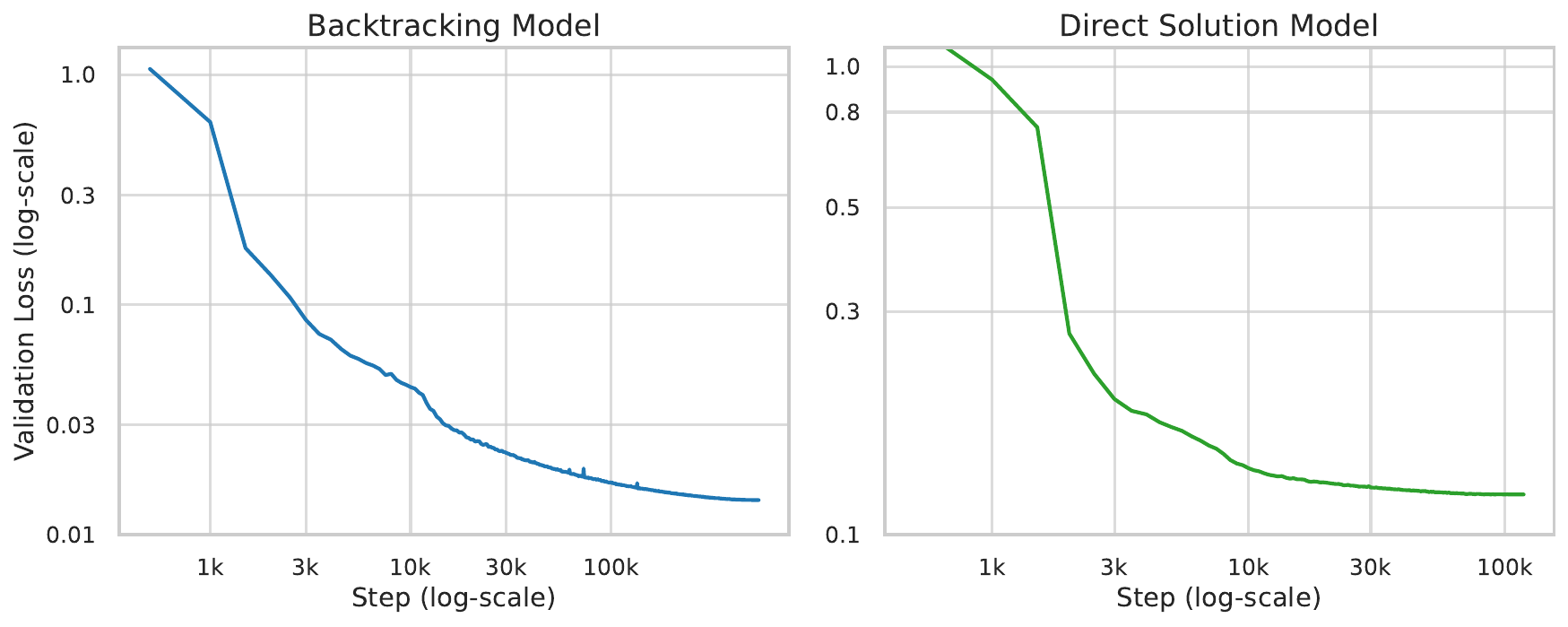}
    \caption{\textbf{CountDown validation loss.}
    \textit{Left: } Backtracking model.
    \textit{Right: } Direct solution model.
    }
    \label{fig:cd_val_loss}

    \includegraphics[width=0.7\linewidth]{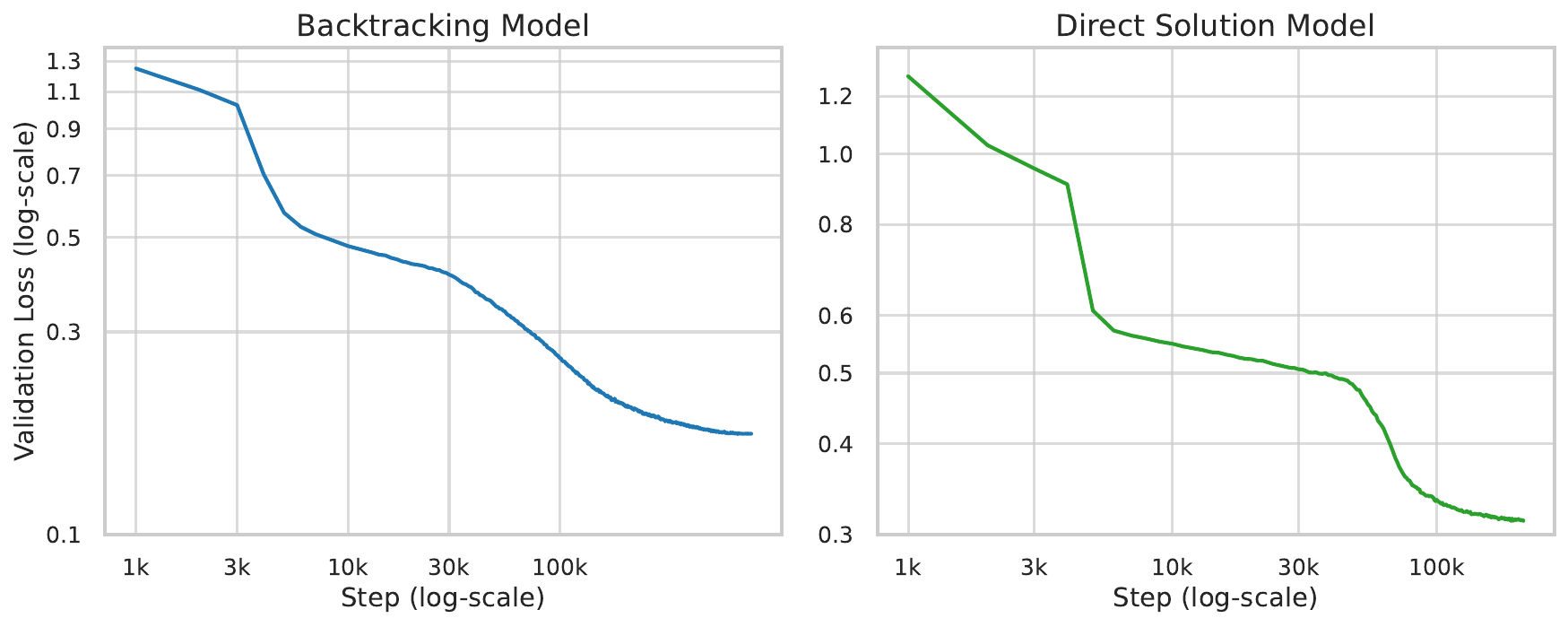}
    \caption{\textbf{Sudoku validation loss.}
    \textit{Left: } Backtracking model.
    \textit{Right: } Direct solution model.
    }
    \label{fig:sudoku_val_loss}
    
\end{figure}

During training, we set the maximum epochs to 30 epochs and allow early stopping. All models converge before 30 epochs and we early stop training when the validation loss has converged on log-log scale. Figure \ref{fig:cd_val_loss}, \ref{fig:sudoku_val_loss} show the training curve for both models and for CountDown and Sudoku. 
\label{appdx:training_curve}

%% file: sections/appendix/grpo.tex
\subsection{Additional GRPO plots}
\label{appdx:rl}
\begin{figure}[ht]
    \centering
    \includegraphics[width=0.8\linewidth]{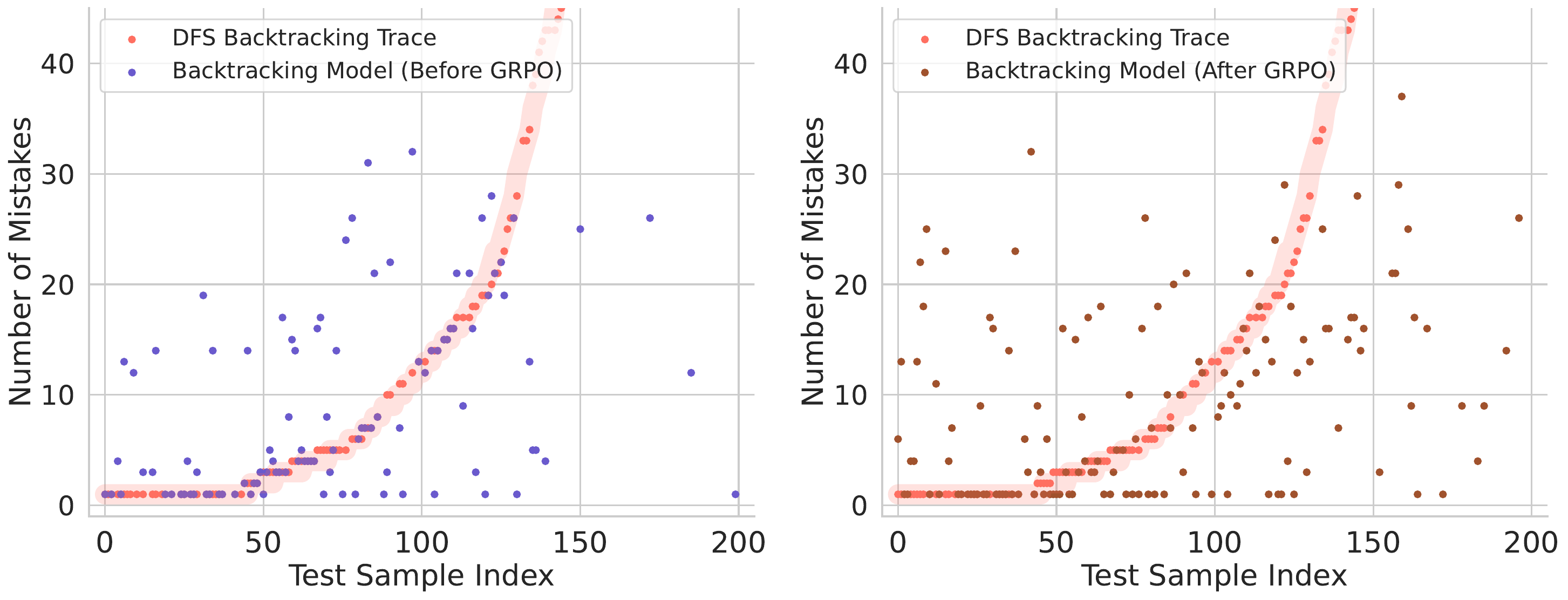}
    \caption{
    \textbf{Backtracking model can discover some new search strategies.} 
    Post GRPO, the backtracking model discover new strategies: In the \textit{right} panel, for each problem, the post-GRPO model makes a different number of mistakes compared to DFS trace.
    }
    \label{fig:grpo_scatter}
\end{figure}

In Figure~\ref{fig:countdown_performance_correlation} (Section~\ref{sec:search_strat}), we used the number of mistakes as a proxy for comparing search strategies. To further demonstrate that the backtracking model fine-tuned with GRPO discovers new strategies, we repeat the same analysis in Figure~\ref{fig:grpo_scatter} (\textit{right}). Compared to the original backtracking model (Figure~\ref{fig:grpo_scatter}, \textit{left}), the post-GRPO model solves many problems with a different number of mistakes than the number of mistakes made by DFS. This shift indicates that the model is no longer tightly aligned with the original search trace and has discovered alternative, more diverse solution paths. Figure \ref{fig:grpo} (\textit{left}) quantifies the above qualitative observation. 

%% file: sections/appendix/verbatim.tex
\section{Data sample}
\label{appx:verbatim}
\begin{figure}[ht]
    \centering
    
    \includegraphics[width=0.8\linewidth]{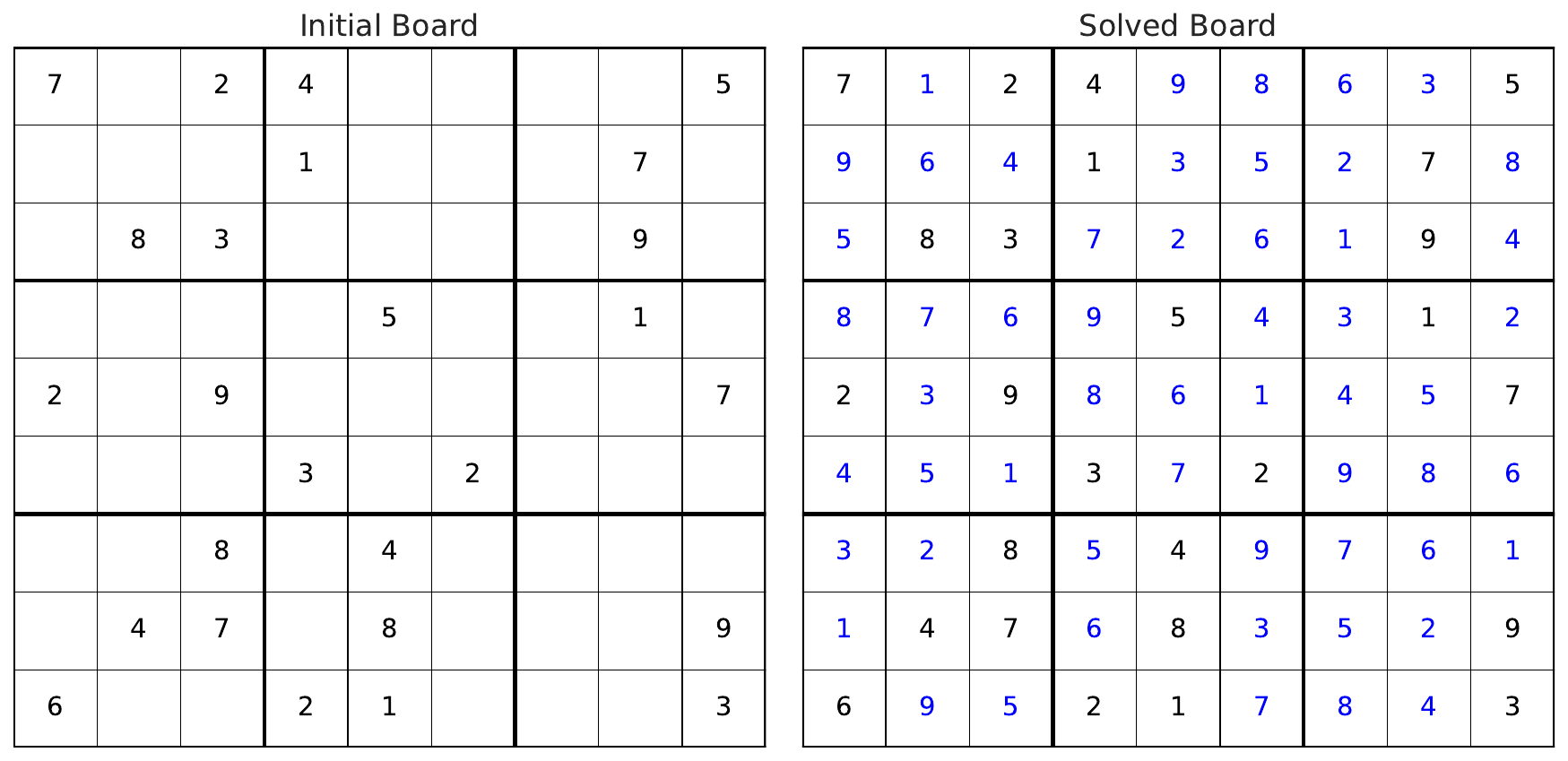}
    \begin{tcolorbox}[colback=yellow!10!white, colframe=black, title=Example of a Sudoku game, fonttitle=\bfseries, boxrule=0.5pt, arc=4pt]
    \small\ttfamily
    START (0, 0) = 7  	(0, 2) = 2  	(0, 3) = 4  	(0, 8) = 5  	(1, 3) = 1  	(1, 7) = 7  	(2, 1) = 8  	(2, 2) = 3  	(2, 7) = 9  	(3, 4) = 5  	(3, 7) = 1  	(4, 0) = 2  	(4, 2) = 9  	(4, 8) = 7  	(5, 3) = 3  	(5, 5) = 2  	(6, 2) = 8  	(6, 4) = 4  	(7, 1) = 4  	(7, 2) = 7  	(7, 4) = 8  	(7, 8) = 9  	(8, 0) = 6  	(8, 3) = 2  	(8, 4) = 1  	(8, 8) = 3  	solving
    \end{tcolorbox}
    
    
    \begin{tcolorbox}[colback=blue!5!white, colframe=black, title=DFS Search Trace, fonttitle=\bfseries, boxrule=0.5pt, arc=4pt]
    \scriptsize\ttfamily
    SOL\_START (4, 4) = 6	(8, 2) = 5	(4, 3) = 8	(8, 1) = 9	(8, 5) = 7	GUESS: (0, 1) [1, 6] = 1	(0, 1) = 1	GUESS: (0, 4) [3, 9] = 3	(0, 4) = 3	GUESS: (0, 6) [6, 8] = 6	(0, 6) = 6	(0, 7) = 8	(8, 7) = 4	(0, 5) = 9	(1, 4) = 2	(1, 8) = 4	(2, 4) = 7	(3, 5) = 4	(4, 5) = 1	(5, 4) = 9	(8, 6) = 8	(1, 2) = 6	(1, 6) = 3	(3, 3) = 7	(1, 1) = 5	(1, 5) = 8	(2, 0) = 4	(4, 1) = 3	(4, 7) = 5	(5, 6) = 4	(5, 7) = 6	(5, 8) = 8	(6, 1) = 2	(7, 7) = 2	(1, 0) = 9	(3, 0) = 8	(3, 1) = 6	(3, 8) = 2	(5, 1) = 7	(5, 2) = 1	(2, 8) = 1	(3, 6) = 9	(5, 0) = 5	(6, 8) = 6	(2, 6) = 2	NO\_CANDIDATE: (3, 2)	NO CANDIDATE: (0, 6)	REVERT: (0, 6) [6, 8] = NONE	GUESS: (0, 6) [6, 8] = 8	(0, 6) = 8	(0, 7) = 6	(8, 6) = 4	(8, 7) = 8	(0, 5) = 9	(1, 4) = 2	(1, 6) = 3	(1, 8) = 4	(2, 4) = 7	(3, 5) = 4	(4, 5) = 1	(4, 6) = 5	(5, 4) = 9	(5, 6) = 6	(5, 7) = 4	(5, 8) = 8	(1, 2) = 6	(3, 3) = 7	(3, 8) = 2	(4, 1) = 3	(5, 2) = 1	(6, 1) = 2	(6, 7) = 5	(7, 7) = 2	(1, 1) = 5	(1, 5) = 8	(2, 0) = 4	(2, 8) = 1	(3, 0) = 8	(3, 1) = 6	(3, 6) = 9	(5, 0) = 5	(5, 1) = 7	(6, 8) = 6	(7, 6) = 1	(1, 0) = 9	(2, 6) = 2	(6, 3) = 9	(6, 5) = 3	(6, 6) = 7	(7, 0) = 3	(6, 0) = 1	NO\_CANDIDATE: (3, 2)	NO\_CANDIDATE: (0, 6)	revert: (0, 6) [6, 8] = NO\_CANDIDATE	NO\_CANDIDATE: (0, 4)	REVERT: (0, 4) [3, 9] = NONE	GUESS: (0, 4) [3, 9] = 9	(0, 4) = 9	(5, 4) = 7	(2, 4) = 2	(3, 3) = 9	(3, 5) = 4	(4, 5) = 1	(1, 4) = 3	(3, 2) = 6	(5, 1) = 5	(1, 1) = 6	(1, 2) = 4	(2, 0) = 5	(2, 5) = 6	(4, 1) = 3	(5, 2) = 1	(6, 1) = 2	(0, 5) = 8	(1, 0) = 9	(1, 5) = 5	(2, 3) = 7	(3, 0) = 8	(3, 1) = 7	(3, 8) = 2	(5, 0) = 4	(7, 5) = 3	(1, 8) = 8	(3, 6) = 3	(5, 8) = 6	(6, 5) = 9	(6, 8) = 1	(7, 0) = 1	(0, 6) = 6	(0, 7) = 3	(1, 6) = 2	(2, 8) = 4	(5, 7) = 8	(6, 0) = 3	(7, 6) = 5	(8, 7) = 4	(2, 6) = 1	(4, 6) = 4	(4, 7) = 5	(5, 6) = 9	(6, 6) = 7	(6, 7) = 6	(7, 3) = 6	(7, 7) = 2	(8, 6) = 8	(6, 3) = 5 SOL END
    \end{tcolorbox}
    
    
    \begin{tcolorbox}[colback=green!10!white, colframe=black, title=Correct Solution, fonttitle=\bfseries, boxrule=0.5pt, arc=4pt]
    \scriptsize\ttfamily
    SOL\_START (4, 4) = 6	(8, 2) = 5	(4, 3) = 8	(8, 1) = 9	(8, 5) = 7	(0, 1) = 1	(0, 4) = 9	(5, 4) = 7	(2, 4) = 2	(3, 3) = 9	(3, 5) = 4	(4, 5) = 1	(1, 4) = 3	(3, 2) = 6	(5, 1) = 5	(1, 1) = 6	(1, 2) = 4	(2, 0) = 5	(2, 5) = 6	(4, 1) = 3	(5, 2) = 1	(6, 1) = 2	(0, 5) = 8	(1, 0) = 9	(1, 5) = 5	(2, 3) = 7	(3, 0) = 8	(3, 1) = 7	(3, 8) = 2	(5, 0) = 4	(7, 5) = 3	(1, 8) = 8	(3, 6) = 3	(5, 8) = 6	(6, 5) = 9	(6, 8) = 1	(7, 0) = 1	(0, 6) = 6	(0, 7) = 3	(1, 6) = 2	(2, 8) = 4	(5, 7) = 8	(6, 0) = 3	(7, 6) = 5	(8, 7) = 4	(2, 6) = 1	(4, 6) = 4	(4, 7) = 5	(5, 6) = 9	(6, 6) = 7	(6, 7) = 6	(7, 3) = 6	(7, 7) = 2	(8, 6) = 8	(6, 3) = 5 SOL\_END
    \end{tcolorbox}
    
    \caption{\textbf{Example of training data for Sudoku game.} \textit{Top}: initial puzzle setup. \textit{Middle}: full search trace with guesses and backtracking (tabs used). \textit{Bottom}: final correct solution.}
    \label{fig:sudoku_data}
\end{figure}

Figure \ref{fig:cd_data} shows an example of a CountDown game and the training data. Figure \ref{fig:sudoku_data} shows an example of a Sudoku game and the training data. Figure \ref{fig:cd_stack_data} shows an example of stacked-CountDown variation and the training data.

\begin{figure}[ht]
\centering
    \begin{tcolorbox}[colback=yellow!10!white, colframe=black, title=Example of a CountDown game, fonttitle=\bfseries, boxrule=0.5pt, arc=4pt]
    \texttt{Target: 16 \hspace{2em} Candidates: [96, 11, 78, 22]}
    \end{tcolorbox}
    
    
    \begin{tcolorbox}[colback=blue!5!white, colframe=black, title=DFS Search Trace, fonttitle=\bfseries, boxrule=0.5pt, arc=4pt]
    \begin{Verbatim}[fontsize=\footnotesize]
    Current State: 16:[96, 11, 78, 22], Operations: []
    Exploring Operation: 96-78=18, Resulting Numbers: [11, 22, 18]
    Generated Node #0,0: 16:[11, 22, 18] Operation: 96-78=18
    Moving to Node #0,0
    Current State: 16:[11, 22, 18], Operations: ['96-78=18']
    Exploring Operation: 22-11=11, Resulting Numbers: [18, 11]
    Generated Node #0,0,0: 16:[18, 11] Operation: 22-11=11
    Moving to Node #0,0,0
    Current State: 16:[18, 11], Operations: ['96-78=18', '22-11=11']
    Exploring Operation: 18-11=7, Resulting Numbers: [7]
    7,16 unequal: No Solution
    Moving to Node #0,0,0
    Current State: 16:[18, 11], Operations: ['96-78=18', '22-11=11']
    Exploring Operation: 18+11=29, Resulting Numbers: [29]
    29,16 unequal: No Solution
    Moving to Node #0,0
    Current State: 16:[11, 22, 18], Operations: ['96-78=18']
    Exploring Operation: 18-11=7, Resulting Numbers: [22, 7]
    Generated Node #0,0,1: 16:[22, 7] Operation: 18-11=7
    Moving to Node #0,0,1
    Current State: 16:[22, 7], Operations: ['96-78=18', '18-11=7']
    Exploring Operation: 22-7=15, Resulting Numbers: [15]
    15,16 unequal: No Solution
    Moving to Node #0,0,1
    Current State: 16:[22, 7], Operations: ['96-78=18', '18-11=7']
    Exploring Operation: 22+7=29, Resulting Numbers: [29]
    29,16 unequal: No Solution
    Moving to Node #0,0
    Current State: 16:[11, 22, 18], Operations: ['96-78=18']
    Exploring Operation: 22/11=2, Resulting Numbers: [18, 2]
    Generated Node #0,0,2: 16:[18, 2] Operation: 22/11=2
    Moving to Node #0,0,2
    Current State: 16:[18, 2], Operations: ['96-78=18', '22/11=2']
    Exploring Operation: 18-2=16, Resulting Numbers: [16]
    16,16 equal: Goal Reached
    \end{Verbatim}
    \end{tcolorbox}
    
    
    \begin{tcolorbox}[colback=green!10!white, colframe=black, title=Direct Solution Trace, fonttitle=\bfseries, boxrule=0.5pt, arc=4pt]
    \begin{Verbatim}[fontsize=\footnotesize]
    Current State: 16:[96, 11, 78, 22], Operations: []
    Exploring Operation: 96-78=18, Resulting Numbers: [11, 22, 18]
    Generated Node #2: [11, 22, 18] from Operation: 96-78=18
    Current State: 16:[11, 22, 18], Operations: ['96-78=18']
    Exploring Operation: 22/11=2, Resulting Numbers: [18, 2]
    Generated Node #3: [18, 2] from Operation: 22/11=2
    Current State: 16:[18, 2], Operations: ['96-78=18', '22/11=2']
    Exploring Operation: 18-2=16, Resulting Numbers: [16]
    16,16 equal: Goal Reached
    \end{Verbatim}
    \end{tcolorbox}
    \caption{\textbf{Example of training data for CoutnDown.} \textit{Top}: game setup. \textit{Middle}: full DFS search trace for training backtracking model. \textit{Bottom}: correct solution path for training direct solution model.}
    \label{fig:cd_data}
\end{figure}

\begin{figure}[ht]
\centering
    \begin{tcolorbox}[colback=yellow!10!white, colframe=black, title=Example of a stacked-CountDown game, fonttitle=\bfseries, boxrule=0.5pt, arc=4pt]
    \texttt{Target: 96 \hspace{2em} Candidates: [22, 77, 24, 48, 31, 12, 36, 35]}
    \end{tcolorbox}
    
    
    \begin{tcolorbox}[colback=blue!5!white, colframe=black, title=DFS Search Trace, fonttitle=\bfseries, boxrule=0.5pt, arc=4pt]
    \begin{Verbatim}[fontsize=\footnotesize]
    Current State: 96:[22, 77, 24, 48, 31, 12, 36, 35], Operations: []
    Exploring Operation: 77-48=29, Resulting Numbers: [22, 24, 29, 31, 12, 36, 35]
    Generated Node #0,0: 96:[22, 24, 29, 31, 12, 36, 35] Operation: 77-48=29
    Moving to Node #0,0
    Current State: 96:[22, 24, 29, 31, 12, 36, 35], Operations: ['77-48=29']
    Exploring Operation: 22+24=46, Resulting Numbers: [29, 46, 31, 12, 36, 35]
    Generated Node #0,0,0: 96:[29, 46, 31, 12, 36, 35] Operation: 22+24=46
    Moving to Node #0,0,0
    Current State: 96:[29, 46, 31, 12, 36, 35], Operations: ['77-48=29', '22+24=46']
    Exploring Operation: 46-29=17, Resulting Numbers: [17, 31, 12, 36, 35]
    17,31 unequal
    Moving to Node #0,0
    ...
    ...
    Current State: 96:[29, 2, 31, 12, 36, 35], Operations: ['77-48=29', '24-22=2']
    Exploring Operation: 29+2=31, Resulting Numbers: [31, 31, 12, 36, 35]
    31,31 equal
    Current State: 96:[31, 12, 36, 35], Operations: []
    Exploring Operation: 36-35=1, Resulting Numbers: [31, 12, 1]
    Generated Node #0,0: 96:[31, 12, 1] Operation: 36-35=1
    Moving to Node #0,0
    Current State: 96:[31, 12, 1], Operations: ['36-35=1']
    Exploring Operation: 31+1=32, Resulting Numbers: [12, 32]
    Generated Node #0,0,0: 96:[12, 32] Operation: 31+1=32
    Moving to Node #0,0,0
    Current State: 96:[12, 32], Operations: ['36-35=1', '31+1=32']
    Exploring Operation: 12+32=44, Resulting Numbers: [44]
    44,96 unequal: No Solution
    ...
    ...
    Exploring Operation: 4*24=96, Resulting Numbers: [96]
    96,96 equal: Goal Reached
    \end{Verbatim}
    \end{tcolorbox}
    
    
    \begin{tcolorbox}[colback=green!10!white, colframe=black, title=Direct Solution Trace, fonttitle=\bfseries, boxrule=0.5pt, arc=4pt]
    \begin{Verbatim}[fontsize=\footnotesize]
    Current State: 96:[22, 77, 24, 48, 31, 12, 36, 35], Operations: []
    Exploring Operation: 77-22=55, Resulting Numbers: [55, 24, 48, 31, 12, 36, 35]
    Generated Node #2: [55, 24, 48, 31, 12, 36, 35] from Operation: 77-22=55
    Current State: 96:[55, 24, 48, 31, 12, 36, 35], Operations: ['77-22=55']
    Exploring Operation: 48-24=24, Resulting Numbers: [55, 24, 31, 12, 36, 35]
    Generated Node #3: [55, 24, 31, 12, 36, 35] from Operation: 48-24=24
    Current State: 96:[55, 24, 31, 12, 36, 35], Operations: ['77-22=55', '48-24=24']
    Exploring Operation: 55-24=31, Resulting Numbers: [31, 31, 12, 36, 35]
    31,31 equal
    Current State: 96:[31, 12, 36, 35], Operations: []
    Exploring Operation: 35-31=4, Resulting Numbers: [4, 12, 36]
    Generated Node #2: [4, 12, 36] from Operation: 35-31=4
    Current State: 96:[4, 12, 36], Operations: ['35-31=4']
    Exploring Operation: 36-12=24, Resulting Numbers: [24, 4]
    Generated Node #3: [24, 4] from Operation: 36-12=24
    Current State: 96:[24, 4], Operations: ['35-31=4', '36-12=24']
    Exploring Operation: 4*24=96, Resulting Numbers: [96]
    96,96 equal: Goal Reached
    \end{Verbatim}
    \end{tcolorbox}
    \caption{\textbf{Example of training data for stacked-CoutnDown} (Appendix \ref{appdx:game_type}). \textit{Top}: game setup. \textit{Middle}: full DFS search trace for training backtracking model. \textit{Bottom}: correct solution path for training direct solution model.}
    \label{fig:cd_stack_data}
\end{figure}